\documentclass[lettersize,journal]{IEEEtran}
\usepackage{cite}
\usepackage{url}

\usepackage{hyperref}
\hypersetup{
    colorlinks=true,
    linkcolor=magenta,
    filecolor=blue,      
    urlcolor=cyan,
    citecolor=green,
}

\usepackage{enumitem}

\usepackage[automake, acronym, toc]{glossaries-extra}
\makeglossaries
\newacronym{rs}{RS}{Remote Sensing}
\newacronym{fms}{FMs}{Foundation Models}
\newacronym{rsfms}{RSFMs}{Remote Sensing Foundation Models}
\newacronym{vlms}{VLMs}{Vision-Language Models}
\newacronym{vfms}{VFMs}{Visual Foundation Models}
\newacronym{llms}{LLMs}{Large Language Models}
\newacronym{eo}{EO}{Earth Observation}
\newacronym{peft}{PEFT}{Parameter-Efficient Tuning}
\newacronym{hsi}{HSI}{HyperSpectral Images}
\newacronym{msi}{MSI}{MultiSpectral Images}
\newacronym{nir}{NIR}{Near-Infrared}
\newacronym{lidar}{LiDAR}{Light Detection and Ranging}
\newacronym{sar}{SAR}{Synthetic Aperture Radar}
\newacronym{ssl}{SSL}{Self-Supervised Learning}
\newacronym{sam}{SAM}{Segment Anything Model}
\newacronym{cv}{CV}{Computer Vision}
\newacronym{gsd}{GSD}{Ground Sample Distances}
\newacronym{mim}{MIM}{Masked Image Modeling}
\newacronym{lulc}{LULC}{Land Use and Land Cover}
\newacronym{uav}{UAVs}{Unmanned Aerial Vehicles}
\newacronym{dsm}{DSM}{Digital Surface Models}
\newacronym{vqa}{VQA}{Visual Question Answering}

\usepackage{pifont}% http://ctan.org/pkg/pifont

\usepackage{fontawesome} % symbols like "download"
\usepackage{subcaption}
\usepackage{multirow}
\usepackage{amsfonts}
\usepackage{adjustbox}
\usepackage{bbding}  % symbols like "envolope"

\begin{document}

\title{Foundation Models for Remote Sensing and Earth Observation:\\A Survey}


\author{Aoran Xiao,
        Weihao Xuan,
        Junjue Wang,
        Jiaxing Huang,\\
        Dacheng Tao~\IEEEmembership{Fellow,~IEEE},
        Shijian Lu, %\textsuperscript{\Envelope},
        and Naoto Yokoya%\textsuperscript{\Envelope}
        ~\IEEEmembership{Member,~IEEE}
\IEEEcompsocitemizethanks{
    \IEEEcompsocthanksitem Aoran Xiao is with the RIKEN Center for Advanced Intelligence Project, Japan.
    \IEEEcompsocthanksitem Weihao Xuan and Naoto Yokoya are with the University of Tokyo and the RIKEN Center for Advanced Intelligence Project, Japan.
    \IEEEcompsocthanksitem Junjue Wang is with the University of Tokyo, Japan.
    \IEEEcompsocthanksitem Jiaxing Huang, Dacheng Tao, and Shijian Lu are with the Nanyang Technological University, Singapore.
    \IEEEcompsocthanksitem Corresponding authors: Naoto Yokoya (naoto.yokoya@riken.jp); Shijian~Lu (shijian.lu@ntu.edu.sg)
    }
}

\markboth{IEEE Geoscience and Remote Sensing Magazine}%
{Shell \MakeLowercase{\textit{et al.}}: A Sample Article Using IEEEtran.cls for IEEE Journals}


\maketitle

\begin{abstract}
Remote Sensing (RS) is a crucial technology for observing, monitoring, and interpreting our planet, with broad applications across geoscience, economics, humanitarian fields, etc. While artificial intelligence (AI), particularly deep learning, has achieved significant advances in RS, unique challenges persist in developing more intelligent RS systems, including the complexity of Earth's environments, diverse sensor modalities, distinctive feature patterns, varying spatial and spectral resolutions, and temporal dynamics. Meanwhile, recent breakthroughs in large Foundation Models (FMs) have expanded AI’s potential across many domains due to their exceptional generalizability and zero-shot transfer capabilities. However, their success has largely been confined to natural data like images and video, with degraded performance and even failures for RS data of various non-optical modalities. This has inspired growing interest in developing Remote Sensing Foundation Models (RSFMs) to address the complex demands of Earth Observation (EO) tasks, spanning the surface, atmosphere, and oceans. This survey systematically reviews the emerging field of RSFMs. It begins with an outline of their motivation and background, followed by an introduction of their foundational concepts. It then categorizes and reviews existing RSFM studies including their datasets and technical contributions across Visual Foundation Models (VFMs), Visual-Language Models (VLMs), Large Language Models (LLMs), and beyond. In addition, we benchmark these models against publicly available datasets, discuss existing challenges, and propose future research directions in this rapidly evolving field. % A project associated with this survey has been built at \url{https://github.com/xiaoaoran/awesome-RSFMs}
\end{abstract}


\section{Introduction}

\IEEEPARstart{T}{he} rapid evolution of deep learning has brought significant advancements to 
\acrfull{rs} and various \acrfull{eo} applications. However, most current models rely on explicitly designed, task-specific learning objectives. This approach demands considerable human effort for dataset collection and annotation, along with substantial computational resources for model training and evaluation. Furthermore, these models exhibit limited generalization and transfer capabilities across different tasks, restricting the broader adoption of RS systems. RS data, sourced from diverse sensors and platforms, is inherently large-scale, complex, dynamic, and heterogeneous. Accurately and intelligently interpreting RS data in a synergistic, robust, and versatile manner remains a critical, yet underexplored, challenge for advancing RS interpretation systems.

\begin{figure}[t]
    \centering
    \includegraphics[width=\linewidth]{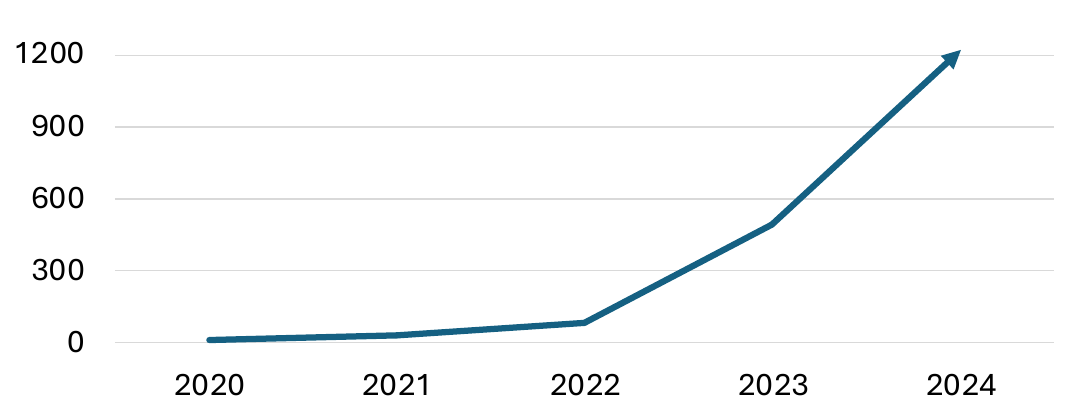}
    \caption{Cumulative number of Google Scholar papers containing the keyphrases 'foundation model' and 'remote sensing' (2020 onward).}
    \label{fig:statistic}
\end{figure}

As deep learning continues to advance, a revolutionary trend has emerged toward large \acrfull{fms}, defined as “\textit{any model trained on broad data (typically using self-supervision at scale) that can be adapted (e.g., fine-tuned) to a wide range of downstream tasks}” \cite{bommasani2021opportunities}. FMs, including \acrfull{llms}, \acrfull{vfms}, and \acrfull{vlms}, have demonstrated remarkable generalization and few-shot transfer capabilities across diverse tasks. This shift marks a transition from single-purpose models to general-purpose models, and from supervised pre-training to self-supervised pre-training, significantly reducing training resource requirements while expanding model application scopes.

\begin{figure*}[t]
    \centering
    \includegraphics[width=\linewidth]{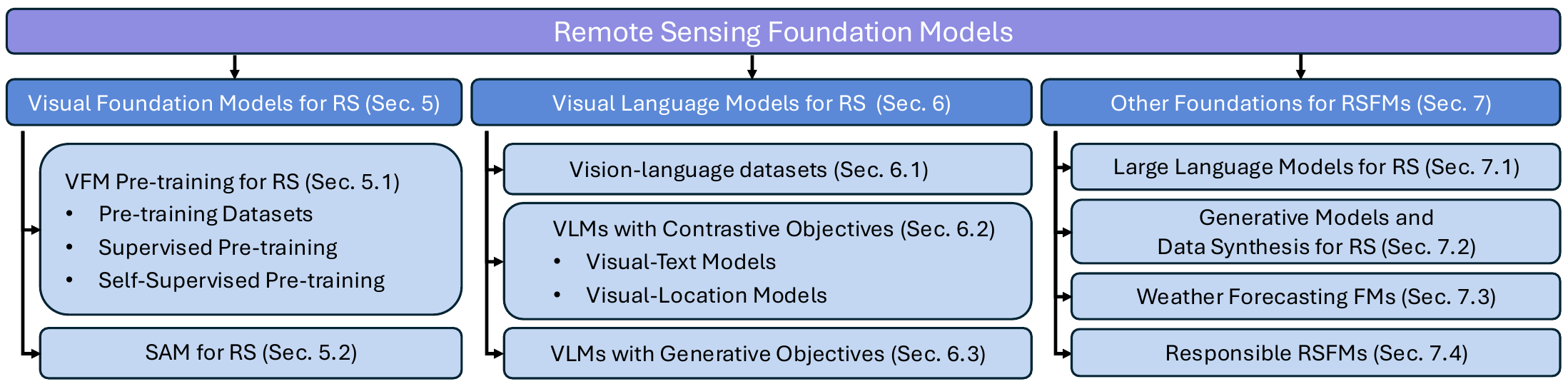}
    \caption{Taxonomy of Remote Sensing Foundation Models.}
    \label{fig:taxonomy}
\end{figure*}

However, these advancements have primarily centered on natural data domains, such as images and texts, which often face significant challenges when applied to out-of-distribution domains like RS. For example, the fundamental differences between RS and natural images—such as sensor modalities, capturing perspectives, spatial resolutions, spectral bands, and temporal regularity—pose obstacles to directly applying FMs in RS applications. Despite these challenges, the success of FMs in natural domains offers promising insights for the development of \acrfull{rsfms}, which have shown great potential in utilizing large-scale geospatial data, modeling complex and dynamic Earth surfaces, improving data efficiency, expanding the range of applications, enhancing task performance, and reducing carbon footprints.

Developing RSFM presents several key challenges compared to FMs in general domains: (1) Significant \textit{domain discrepancies} between natural and RS data; (2) A shortage of \textit{massive datasets} for RSFM pre-training; (3) The absence of suitable \textit{deep architectures} tailored for RSFMs; and (4) The need to address \textit{unique RS applications} that differ from general-purpose FMs in natural domains. To tackle these challenges, increasing efforts have recently focused on developing advanced RSFMs and better integrating various FMs within the RS domain, as illustrated in Fig. \ref{fig:statistic}.

Despite rapid progress, the field of RSFMs still lacks a comprehensive survey that provides an in-depth overview of this emerging and multifaceted area. This paper aims to bridge this gap by presenting an extensive survey of the latest advancements in RSFMs. We explore the field from various perspectives, including learning paradigms, datasets, technical approaches, benchmarks, and future research directions. As illustrated in Fig. \ref{fig:taxonomy}, we categorize existing methods into three main groups based on their \textit{model types}: VFMs for RS, VLMs for RS, and other RSFMs such as LLMs and generative FMs. These categories will be reviewed in detail in the subsequent sections.

\begin{figure}[t]
    \centering
    \includegraphics[width=\linewidth]{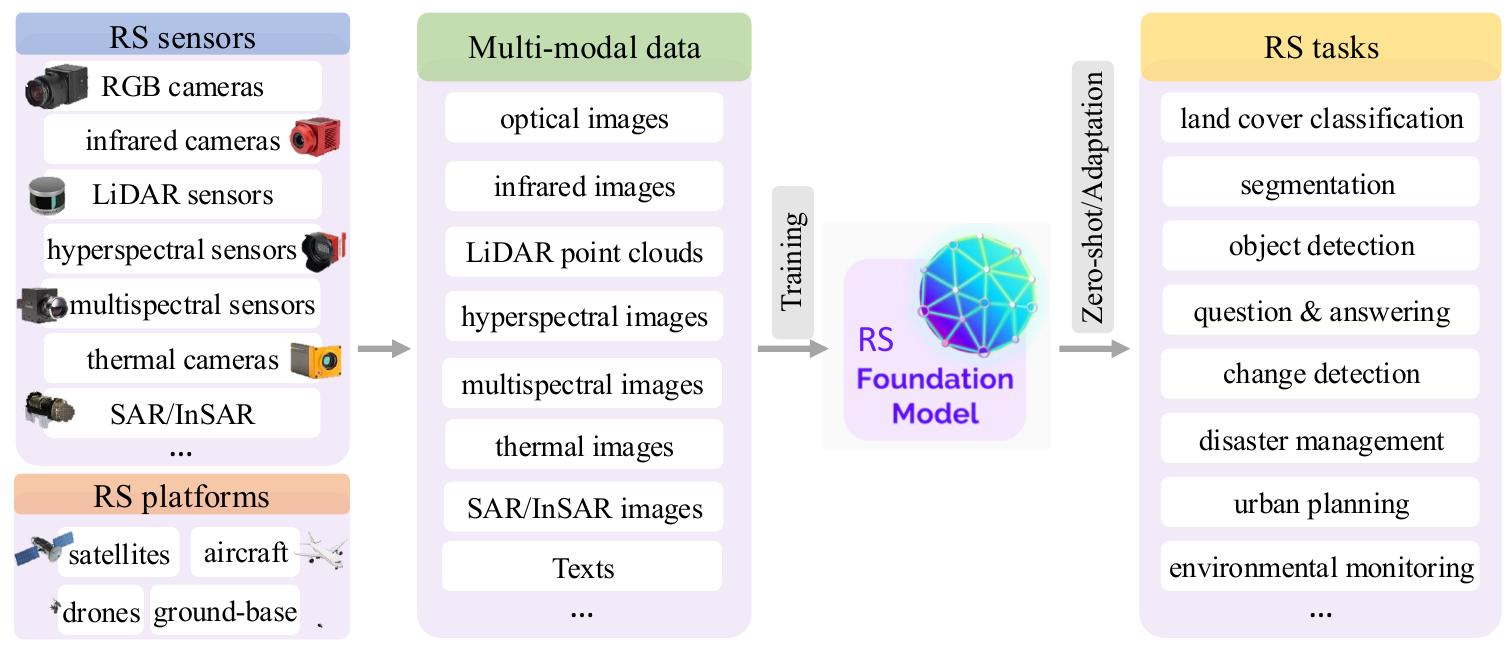}
    \caption{Overview of RSFMs.}
    \label{fig:Overview}
\end{figure}
    
The major contributions of this work are threefold: \textit{First}, it provides a thorough and systematic review of the latest advancements in RSFMs. \textit{Second}, it benchmarks and offers an in-depth analysis of RSFMs applied across various sensor modalities and tasks. \textit{Third}, it identifies several research challenges and proposes potential research directions in the domain of RSFMs.

The structure of this survey is as follows: In Section \ref{sec.background}, we provide background knowledge on RSFMs, including learning paradigms, common RS sensor modalities, and related surveys. Section \ref{sec.foundations} delves into the foundations of RSFMs, covering deep network architectures and typical RS interpretation tasks. Sections \ref{sec:vfm}, \ref{sec:vlm}, and \ref{sec:other_fms} offer a systematic review of methods for VFMs in RS, VLMs in RS, and other types of RSFMs. In Section \ref{sec:benchmarks}, we summarize and compare the performance of existing methods across multiple benchmark datasets. Finally, Section \ref{sec.future} outlines several promising future directions for RSFMs.

\section{Background}\label{sec.background}

\subsection{RS Learning Paradigms}

This subsection briefly outlines the evolution of learning paradigms in RS models, from traditional machine learning, deep learning, culminating in the current FM paradigm. In the following, we provide a concise introduction to each paradigm, highlighting their key differences and advancements, as well as their impact on RS tasks.

\noindent \textbf{(1) Traditional Machine Learning.} In this paradigm, RS relied on manually crafted features and simple learning models that categorized these features into predefined classes. However, this approach depended heavily on domain expertise for feature creation, making it less effective for complex tasks and scenarios within RS domains. Consequently, its scalability and generalizability were significantly limited.

\noindent \textbf{(2) Deep Learning from Scratch and Prediction.} Deep learning revolutionized RS interpretations by replacing complex feature engineering with end-to-end trainable deep neural networks (DNNs), significantly improving accuracy and robustness of RS models. Research of this paradigm emphasized DNN architecture design to extract effective features from various RS sensor modalities across \acrshort{eo} tasks. However, several challenges still remain: 1) RS DNNs are tailored for specific tasks, limiting their generalizability; 2)~Training from scratch leads to slow convergence; and 3)~Collecting and annotating large-scale training data is labor-intensive, time-consuming, and costly.

\noindent \textbf{(3) FM Learning paradigm.} The learning of FM typically involves two primary stages as shown in Fig.~\ref{fig:Overview} : 1) \textit{Pre-training}, where the model learns generalizable and transferable representations, and 2) \textit{Utilization}, where the pre-trained model is applied to downstream tasks.

The \textit{pre-training} stage can be divided into two common approaches:
\begin{itemize}[noitemsep,nolistsep,topsep=0pt,parsep=0pt,partopsep=0pt]
    \item \textbf{Supervised Pre-training.} This approach involves pre-training DNNs on large-scale \textit{labeled} datasets (e.g., ImageNet \cite{deng2009imagenet}) using supervised loss objectives. While it achieves state-of-the-art performance in many downstream tasks, this method requires extensive labeled data, which can be costly to collect.
    \item \textbf{Unsupervised Pre-training.} This approach utilizes self-supervised learning \cite{jing2020self,xiao2023unsupervised} to learn useful and transferable representations from \textit{unlabeled} data by optimizing various unsupervised pre-text tasks. This approach is particularly advantageous in the RS domain, where numerous sensors on different platforms like satellites continuously capture vast amounts of data that are almost impractical to annotate. On the other hand, the pre-trained model may not be directly applicable to specific tasks.
\end{itemize}

Following the pre-training stage, the \textit{utilization} of FMs can be conducted through three common approaches:
\begin{itemize}[noitemsep,nolistsep,topsep=0pt,parsep=0pt,partopsep=0pt]
    \item \textbf{Fully Fine-tuning and Prediction.} This approach utilizes the strong representations of FMs as a starting point, fully fine-tuning all model parameters for specific downstream tasks to aid convergence and boost performance. However, it overwrites and loses the original representations of the powerful FMs.
    \item \textbf{Parameter-efficient Tuning and Prediction.} Unlike fully fine-tuning, \acrfull{peft} introduces only lightweight, learnable parameters while keeping the FM backbone frozen. This allows for the efficient learning of domain-specific or task-specific features while preserving the FM's powerful representation space. PEFT is particularly beneficial in scenarios (such as RS) where FMs face performance challenges due to data distribution gaps and diverse task objectives, enabling adaptation to these domains and tasks without compromising the FM's capabilities.
    \item  \textbf{Zero-shot Prediction.} FMs trained on large-scale data often demonstrate strong zero-shot prediction capabilities, making predictions without the need for domain- or task-specific fine-tuning. However, due to significant domain discrepancies between natural and RS data, FMs of general domains trained on natural datasets frequently underperform in RS scenarios. Additionally, the lack of web-scale RS pre-training datasets means that, regrettably, no RSFM currently exhibits as robust zero-shot capabilities as FMs in general domains.
\end{itemize}

\subsection{Common RS Sensor Modalities}

\begin{figure}[t]
    \centering
    \includegraphics[width=\linewidth]{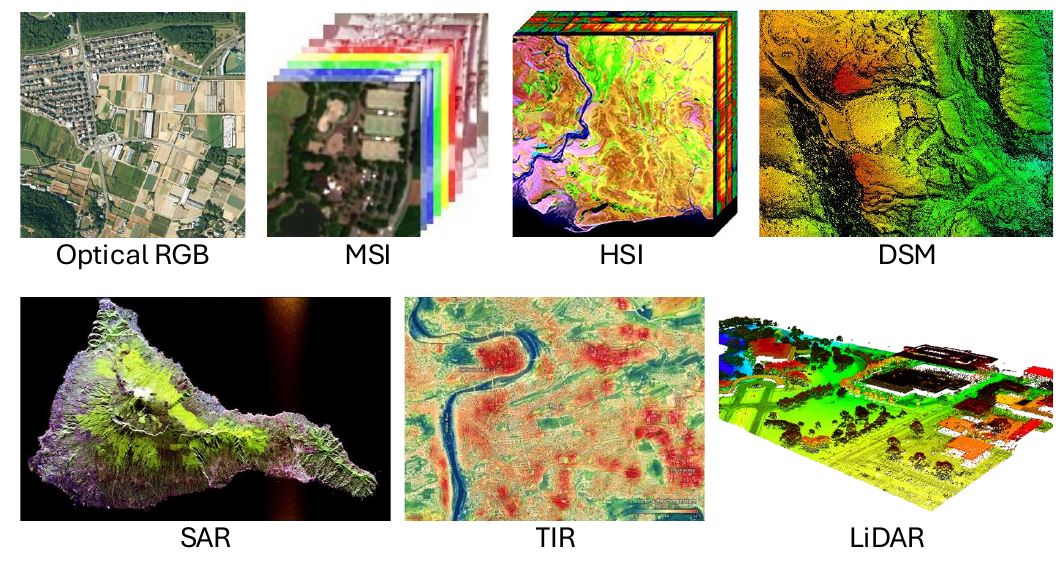}
    \caption{Example of different RS modalities.}
    \label{fig:rs-modalities}
\end{figure}

This subsection provides an overview of the RS sensor modalities commonly employed in existing RSFMs. Examples of these modalities are illustrated in Fig.~\ref{fig:rs-modalities}.
\begin{itemize}[noitemsep,nolistsep,topsep=0pt,parsep=0pt,partopsep=0pt]
    \item \textbf{Optical RGB images}, or true-color images, are among the most widely utilized sensor modalities in RS. They capture visible light in the red, green, and blue spectral bands via cameras deployed on platforms like satellites, aircraft,  unmanned aerial vehicles, and ground-based vehicles. While general FMs can be directly applied to these images, performance is often suboptimal due to domain discrepancies between RS and natural images.
    \item \textbf{\acrfull{msi}} capture data across multiple spectral bands, extending beyond the visible RGB range to include portions of the \acrfull{nir} and shortwave infrared (SWIR) regions. While these images provide more spectral information than RGB, they face challenges when applied to general foundation models due to input incompatibilities, domain gaps, and the increased complexity of spectral data.
    \item \textbf{\acrfull{hsi}} capture data across tens of narrow, contiguous spectral bands, offering highly detailed spectral information compared to MSI and optical RGB images. This fine spectral resolution allows for precise identification of materials and substances, making hyperspectral sensors ideal for tasks such as mineral exploration, vegetation analysis, and environmental monitoring. However, the high dimensionality of HSI introduces challenges such as computational complexity and the risk of overfitting. Additionally, it is not easily compatible with general FMs due to the unique spectral characteristics and domain gaps.
    \item \textbf{\acrfull{sar}} uses active microwave signals to capture images, enabling data collection in all weather conditions, day and night. SAR provides detailed surface information, revealing insights into surface structure and material properties, making it particularly valuable for applications such as terrain mapping, disaster monitoring, and forest structure analysis. 
    Polarimetric SAR (PolSAR) further enhances SAR's capabilities by measuring the polarization of radar waves, providing deeper understanding of surface characteristics.
    Interferometric SAR (InSAR) works by combining multiple SAR images of the same area to detect minute surface deformations through phase difference analysis, which is crucial for precise terrain mapping. However, SAR images exhibit unique geometric characteristics due to slant-range projection, differing from traditional central projection systems. Furthermore, their complex natures, including speckle noise, present significant challenges for general FMs.
    \item \textbf{\acrfull{lidar} point clouds} capture three-dimensional (3D) spatial data by emitting laser pulses and measuring the time it takes for them to return after reflecting off surfaces. This results in highly accurate 3D representations of terrain and objects, making LiDAR ideal for tasks such as topographic mapping, forest structure analysis, and urban modeling. Despite the rich geometric information, the irregular and sparse nature of point clouds presents challenges for applying general FMs for natural data like images.
    \item \textbf{Thermal Infrared (TIR) Images} capture the heat emitted by objects providing data based on temperature differences. This enables TIR imaging to be useful for applications like environmental monitoring, urban heat island analysis, vegetation health assessment, and wildfire detection. TIR sensors detect emitted radiation in the infrared spectrum, allowing for thermal mapping even in low-light or nighttime conditions. However, the spatial resolution of TIR images is often lower compared to optical sensors, and the data is affected by atmospheric conditions and surface emissivity, posing challenges for accurately interpreting thermal information in general FMs.
    \item \textbf{\acrfull{dsm}}  represent the elevation of the Earth's surface, including all objects such as buildings, vegetation, and infrastructure. DSMs are typically derived from LiDAR, radar, or stereo imagery and are widely used in urban planning, flood modeling, and landscape analysis. While DSMs provide valuable 3D information about surface features, their grid-like structure and elevation-specific data pose challenges for general FMs.
\end{itemize}

\begin{table}[t]
    \centering
    \caption{A summary of related surveys on remote sensing foundation models.}
    \label{tab:surveys}
    \begin{tabular}{|l|c|l|}
    \hline
       Pub. & Year & Scopes \& Contributions \\
    \hline
       \cite{wang2022self} & 2022 & Developments in \acrshort{ssl} for \acrshort{cv} in the context of \acrshort{rs}.\\
       \cite{jiao2023brain} & 2023 & Developments of \acrshort{ssl} for various RS applications.\\
       \cite{zhu2024foundations} & 2024 & Foundations for earth and climate \acrshort{rsfms}. \\
       \cite{li2024vision} & 2024 & Progress and trends for \acrshort{vlms} in RS. \\
       \cite{hadid2024geoscience} & 2024 & Generative AI and  \acrshort{llms} for RS. \\
       \cite{zhang2024geoscience} & 2024 & Broad overview of FMs in geoscience.\\
       \textbf{Ours} & \textbf{2025} & \textbf{RSFMs in dataset development, model taxonomy,} \\
       & & \textbf{technical advances, and benchmarking.}\\
    \hline
    \end{tabular}
\end{table}

\subsection{Relevant Surveys}

This survey systematically review and benchmark FMs across various RS fields, including VFMs, VLMs, LLMs, generative FMs, and beyond. While numerous surveys in natural domains \cite{bommasani2021opportunities,zhang2024vision} cover diverse scopes and applications (see Table \ref{tab:surveys}), they do not specifically address geoscience and RS modalities. Additionally, there are several surveys regarding in geoscience and RS but they focus on specific topics such as foundations for earth and climate RSFMs \cite{zhu2024foundations}, self-supervised learning \cite{wang2022self,jiao2023brain}, and vision-language modeling \cite{zhou2024towards}. 
\cite{zhang2024geoscience} present present a broad yet high-level review of FMs in geoscience, primarily emphasizing general-domain milestone models rather than geoscience-specific advancements. The review lacks in-depth analysis of dataset resources and technical progress in RS and geoscience, limiting its effectiveness in capturing the latest field developments. Moreover, these reviews predominantly rely on qualitative analyses, omitting the quantitative benchmarking necessary for a more comprehensive evaluation. Our survey aims to bridge these gaps by providing a comprehensive and up-to-date overview of various RSFM types and RS modalities, highlighting the latest advancements in the field.

\begin{figure}[t]
    \centering
    \includegraphics[width=0.78\linewidth]{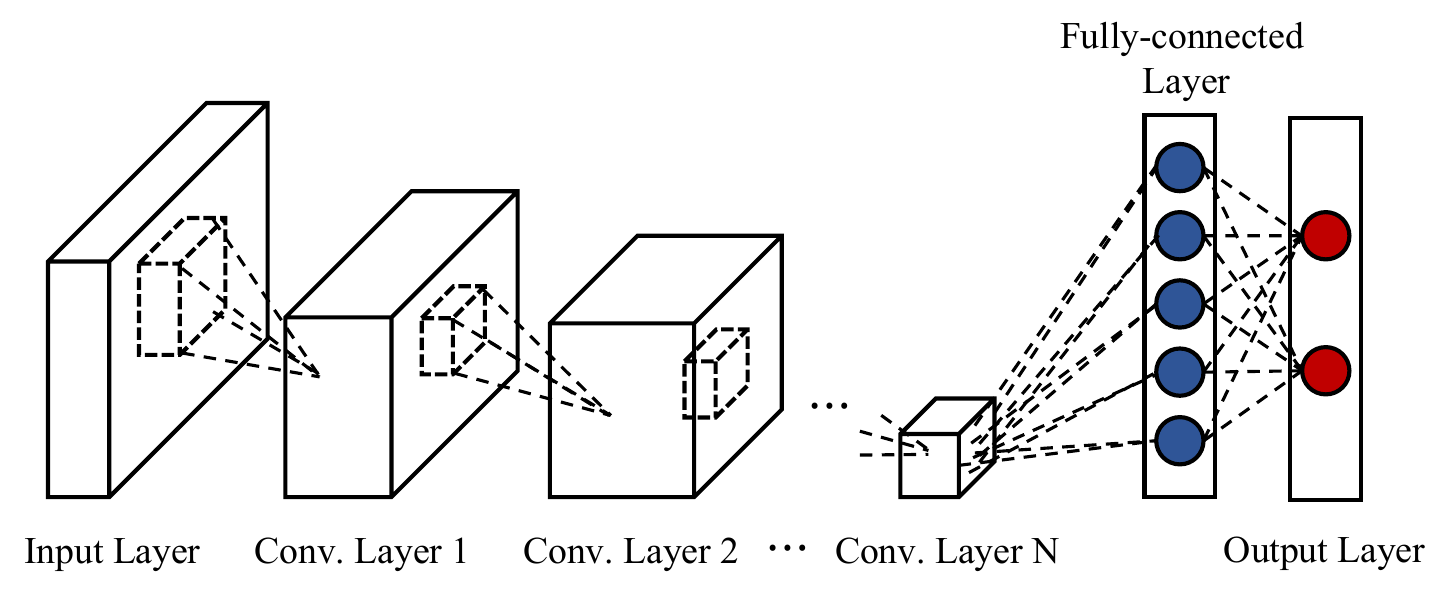}
    \caption{CNN-based deep neural architecture.  Figure is sourced from \cite{zhang2024vision}.}
    \label{fig:cnn}
\end{figure}

\begin{table*}[t]
    \scriptsize
    \centering
    \caption{Summary of commonly used datasets in RSFM pre-training.}
    \label{tab. pre-train-datasets}
    \begin{tabular}{|l|c|c|c|c|c|c|c|}
    \hline
        Dataset & Date & \#Samples & Modal & Annotations & Data Sources & GSD & Link \\
    \hline
        FMoW-RGB\cite{christie2018functional} & 2018 & 363.6k & RGB & 62 classes & QuickBird-2, GeoEye-1, WorldView-2/3 & varying & \href{https://github.com/fMoW/dataset}{\faDownload}\\
        BigEarthNet\cite{sumbul2019bigearthnet} & 2019 & 1.2 million & MSI,SAR & 19 LULC classes & Sentinel-1/2 & 10,20,60m & \href{https://bigearth.net}{\faDownload}\\
        SeCo\cite{manas2021seasonal} & 2021 & 1 million & MSI & None & Sentinel-2; NAIP & 10,20,60m & \href{https://github.com/ServiceNow/seasonal-contrast?tab=readme-ov-file}{\faDownload}\\
        FMoW-Sentinel\cite{cong2022satmae} & 2022 & 882,779 & MSI & None & Sentinel-2 & 10m & \href{https://sustainlab-group.github.io/SatMAE/}{\faDownload}\\
        MillionAID\cite{wang2022empirical} & 2022 & 1 million & RGB & 51 LULC classes & SPOT, IKONOS,WorldView, Landsat, etc. & 0.5m-153m & \href{https://captain-whu.github.io/DiRS/}{\faDownload}\\
        GeoPile\cite{mendieta2023towards} & 2023 & 600K & RGB & None & Sentinel-2, NAIP, etc. & 0.1m-30m & \href{https://github.com/mmendiet/GFM}{\faDownload}\\
        SSL4EO-S12\cite{wang2023ssl4eo} & 2023 & 3 million & MSI, SAR & None & Sentinel-1/2 & 10m & \href{https://github.com/zhu-xlab/SSL4EO-S12}{\faDownload} \\
        SatlasPretrain\cite{bastani2023satlaspretrain} & 2023 & 856K tiles & RGB,MSI,SAR & 137 classes of 7 types & Sentinel-1/2, NAIP, NOAA Lidar Scans & 0.5–2m,10m & \href{https://github.com/allenai/satlas/blob/main/SatlasPretrain.md}{\faDownload}\\
        MMEarth\cite{nedungadi2024mmearth} & 2024 & 1.2 million & RGB,MSI,SAR,DSM & None & Sentinel-1/2, Aster DEM, etc. & 10,20,60m & \href{https://github.com/vishalned/MMEarth-data}{\faDownload}\\
        HyperGlobal-450K\cite{wang2025hypersigma} & 2025 & 450K & HSI & None & EO-1, GF-5 & 30m & \href{https://huggingface.co/datasets/WHU-Sigma/HyperGlobal-450K/tree/main}{\faDownload} \\
        Hyper-Seg\cite{li2025hyperfree} & 2025 & 41,900  & HSI & Masks & AVIRIS, Sentinel-2, GF-series, etc. & 0.6-5m & \href{https://huggingface.co/JingtaoLi/HyperFree/tree/main}{\faDownload}\\
        
    \hline
    \end{tabular}
\end{table*}

\section{Foundations of RSFMs}\label{sec.foundations}

FMs are underpinned by two fundamental technical elements: \textit{transfer learning} and \textit{scale} \cite{bommasani2021opportunities}. \textit{Transfer learning} involves utilizing knowledge gained from one task or modality to improve performance on another. In deep learning, the primary method for transfer learning is pretraining, where a model is first trained on a surrogate task and then fine-tuned for a specific downstream task.  While transfer learning facilitates the development of FMs, it is \textit{scale} that enhances their power. This scale depends on three critical factors: (i) advancements in computing hardware, particularly GPUs; (ii) the progression of deep learning architectures; and (iii) the availability of large-scale training datasets.

In Sections \ref{sec:vfm}, \ref{sec:vlm} and \ref{sec:other_fms}, we will provide a detailed review of the pretraining methods used for RSFMs, along with their corresponding pre-training datasets. In this section, we will first introduce foundational deep architectures in subsection \ref{subsec:architectures} and then discuss the most common downstream RS tasks in subsection \ref{subsec:rs-tasks}.

\subsection{Deep Network Architectures}\label{subsec:architectures}

\noindent\textbf{Convolutional Neural Networks (CNNs).} CNNs are foundational deep learning models, widely used in visual tasks due to their ability to learn spatial hierarchies of features through localized receptive fields, as illustrated in Fig.~\ref{fig:cnn}. Among these, ResNet \cite{he2016deep} stands out as a prominent architecture, incorporating skip connections to address issues like vanishing or exploding gradients, allowing for the construction of very deep networks. While CNNs are prevalent in many RS applications, their role in RSFMs is primarily focused on pre-training for VFMs.

\begin{figure}[t]
    \centering
    \includegraphics[width=\linewidth]{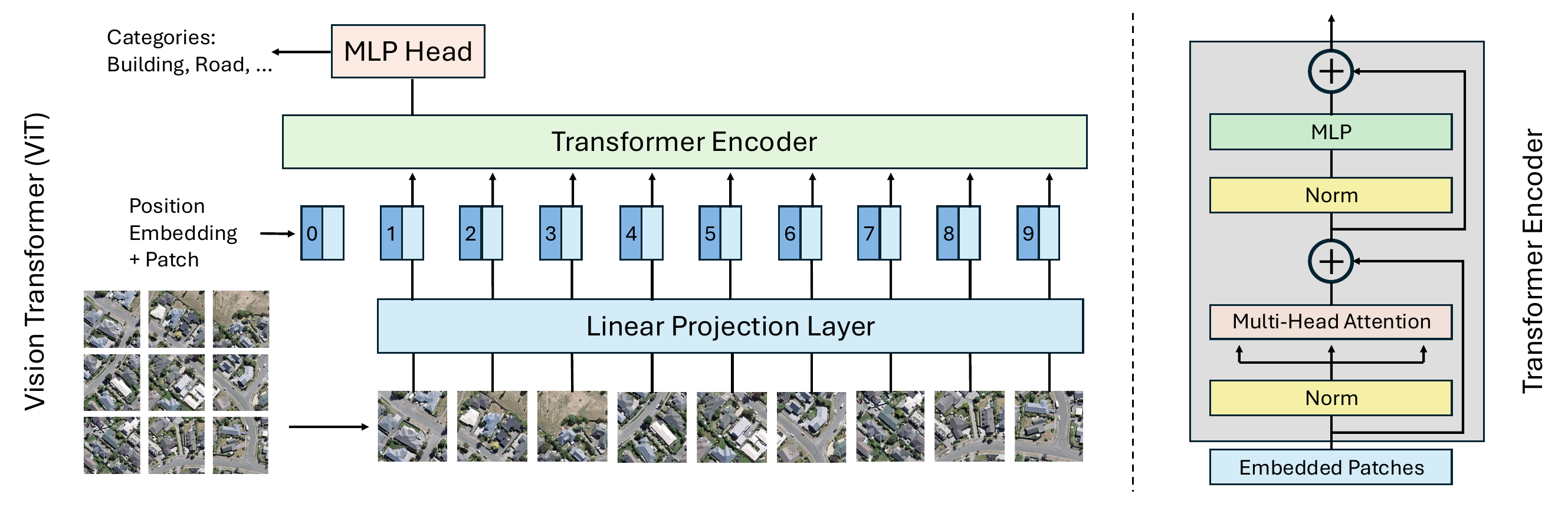}
    \caption{Architecture of Vision Transformer (ViT)~\cite{dosovitskiy2020image}.}
    \label{fig:transformer}
\end{figure}

\noindent\textbf{Transformers.} Transformers~\cite{vaswani2017attention,dosovitskiy2020image}, as depicted in Fig.~\ref{fig:transformer}, are designed to process sequence data using self-attention mechanisms, enabling them to capture relationships between data points regardless of their positional distance. Unlike CNNs, which emphasize local features, transformers excel at modeling global dependencies, making them particularly effective for long-range interactions. A significant advantage of transformers in FMs is their ability to integrate across multiple modalities, enabling them to process a diverse range of inputs, such as visual data (e.g., images, depth, thermal) and non-visual data (e.g., text, 3D point clouds, audio). This cross-modality integration paves the way for unified RSFMs capable of handling a wide spectrum of geospatial data modalities.

\subsection{Typical RS Interpretation Tasks}\label{subsec:rs-tasks}

\begin{itemize}[noitemsep,nolistsep,topsep=0pt,parsep=0pt,partopsep=0pt]
    \item \textbf{Scene Classification} involves categorizing entire image scenes into predefined \acrfull{lulc} classes, such as urban, forest, water, or agricultural areas. While scene classification is a fundamental RS task, it presents challenges due to variations in spatial resolution, spectral differences, and the complexity of natural scenes.
    \item \textbf{Semantic Segmentation} involves classifying each pixel in an image into a specific category, such as water, vegetation, or built-up areas, providing granular insights into LULC at the pixel level. However, semantic segmentation in RS faces unique challenges such as spectral ambiguity, where different objects may share similar spectrum or spectral variation within the same objects. Additionally, the intricate spatial patterns and the need to process multi-sensor and multi-resolution data further complicate the task, as these datasets often differ significantly from those used in natural image segmentation.
    \item \textbf{Object Detection} identifies and locates objects such as buildings, vehicles, or ships within an image by enclosing them in bounding boxes. There are two main types of RS detection: \textit{horizontal}, which uses axis-aligned rectangles, and \textit{arbitrary-oriented}, where the bounding boxes are rotated to match the object’s orientation, addressing the varied angles typical in RS imagery. This task is essential for applications like urban infrastructure analysis and surveillance, but it faces challenges such as scale variations, occlusions.
    \item \textbf{Change Detection} involves identifying differences in an area by comparing images captured at different times, allowing for the detection of changes in land cover, urban development, vegetation health, or disaster impacts. This task plays a critical role in environmental monitoring, tracking urban expansion, and disaster response. However, it faces challenges such as variations in lighting conditions, seasonal changes, sensor inconsistencies, and ensuring precise alignment of multi-temporal images for accurate analysis.
    \item \textbf{\acrfull{vqa}} involves answering language questions based on image content. The goal is to enable users to interact with RS imagery through intuitive questions, such as identifying objects, counting items, or assessing changes in \acrshort{lulc}. VQA is particularly useful for non-expert users seeking insights from complex geospatial data. Key challenges include interpreting the context of the professional question, linking it to relevant visual information across different RS modalities etc.
    \item \textbf{Image Captioning} entails generating descriptive text for images, summarizing key features and content. This task makes complex RS imagery more accessible by providing descriptions of elements such as land cover types, urban structures, or environmental changes. It supports documentation, reporting, and data accessibility. However, challenges arise in producing accurate and detailed captions that are both relevant and informative, given the distinct characteristics of RS data and the limited availability of corresponding textual descriptions compared to natural image domains.
    \item \textbf{Visual Grounding} refers to the task of linking textual queries or descriptions to specific regions or objects within images. In RS, it enables the localization of features like buildings or land cover types based on written descriptions. This task is especially valuable for querying satellite imagery and validating observations against ground truth. Key challenges include managing the spatial and spectral complexity of RS data and accurately aligning text with the relevant visual elements.
\end{itemize}

\section{Visual Foundation Models for RS}\label{sec:vfm}

\begin{table*}[t]
    \scriptsize
    \setlength\tabcolsep{1.5pt}
    \centering
    \caption{VFMs for RS with different Pre-training strategies. "Sup." denotes supervised pre-training; "SSL-C", "SSL-M", and "SSL-C\&M" denote self-supervised contrastive pre-training, self-supervised masked image modelling and their combination, respectively.}
    \label{tab:VFM-pretrain}
    \begin{tabular}{|l|c|c|c|l|}
    \hline
       Model  & Pre-train & Publication & Modal & Contribution \\
    \hline
       RSP\cite{wang2022empirical} \href{https://github.com/ViTAE-Transformer/RSP}{[code]}  & Sup. & TGRS2022 & RGB & Empirical study of supervised classification pretraining with large scale aerial images.\\
       SatlasNet\cite{bastani2023satlaspretrain} \href{https://github.com/allenai/satlas}{[code]} & Sup. & ICCV2023 & RGB,MSI & Multi-task supervised pre-training with large scale aerial images. \\
       SeCo\cite{manas2021seasonal} \href{https://github.com/ServiceNow/seasonal-contrast}{[code]} & SSL-C & ICCV2021 & MSI & Contrast across seasonal changes to learn time and position invariance.\\
       GASSL\cite{ayush2021geography} \href{https://github.com/sustainlab-group/geography-aware-ssl}{[code]} & SSL-C & ICCV2021 & RGB & Pre-training with temporal contrastive learning and geo-location classification.\\
       MATTER\cite{akiva2022self} \href{https://github.com/periakiva/MATTER}{[code]} & SSL-C & CVPR2022 & RGB & Contrast temporally unchanged regions to learn material and texture representations.\\
       CACo\cite{mall2023change} \href{https://github.com/utkarshmall13/CACo}{[code]} & SSL-C & CVPR2023 & RGB & Change-aware sampling and contrastive learning for temporal satellite images.\\
       CSP\cite{mai2023csp} \href{https://github.com/gengchenmai/csp}{[code]} & SSL-C & ICML2023 & RGB & Contrastive spatial pre-training for geo-tagged images.\\
       SkySense\cite{guo2024skysense} & SSL-C & CVPR2024 & RGB, MSI, SAR & Contrast across modals and spatial granularities with geo-context prototype learning.\\ 
       CRISP\cite{huynh2024contrastive} & SSL-C & ECCV2024 & RGB & Contrast between ground-level and aerial image pairs. \\      SatMAE\cite{cong2022satmae}\href{https://github.com/sustainlab-group/SatMAE}{[code]} & SSL-M & NIPS2022 & RGB, MSI & Temporal and multi-spectral MAE.\\
       GFM\cite{mendieta2023towards} \href{https://github.com/mmendiet/GFM}{[code]} & SSL-M & ICCV2023 & RGB & Continual MIM pretraining from imagenet to the geospatial domain.\\
       Scale-MAE\cite{reed2023scale} \href{https://github.com/bair-climate-initiative/scale-mae}{[code]} & SSL-M & ICCV2023 & RGB & MAE with low/high frequency reconstruction and a ground sample distance positional encoding. \\
       msGFM\cite{han2024bridging} \href{https://github.com/boranhan/Geospatial_Foundation_Models}{[code]} & SSL-M & CVPR2024 & RGB,MSI,SAR,DSM & Multimodal MIM with a shared encoder and distinct patch embedding/decoder.\\
       SatMAE++\cite{noman2024rethinking} \href{https://github.com/techmn/satmae_pp}{[code]} & SSL-M & CVPR2024 & RGB,MSI & MAE pre-training with multi-scale reconstructions.\\
       MA3E\cite{li2024masked} \href{https://github.com/benesakitam/MA3E}{[code]} & SSL-M & ECCV2024 & RGB & Pre-training with angle-aware MIM for learning angle-invariant representations. \\
       OmniSat \cite{astruc2024omnisat} \href{https://github.com/gastruc/OmniSat}{[code]} & SSL-C & ECCV2024 & RGB, SAR & Cross-modal contrast among temporal and high-resolution RGB images and SAR images.\\
       MMEarth\cite{nedungadi2024mmearth} \href{https://vishalned.github.io/mmearth/}{[code]} & SSL-M & ECCV2024 & MSI,SAR,DEM,etc. & Multimodal MAE with 12 paired pixel-level and image-level modalities.\\
       CROMA\cite{fuller2024croma} \href{https://github.com/antofuller/CROMA}{[code]} & SSL-C\&M & NIPS2023 & MSI,SAR & Pre-training with cross-modal contrastive learning and multimodal MIM.\\
       CS-MAE\cite{tang2023cross} \href{https://github.com/aicip/Cross-Scale-MAE}{[code]} & SSL-C\&M & NIPS2023 & RGB & Cross-scale pre-training for both contrastive consistency and MIM reconstruction. \\
       HyperSIGMA \cite{wang2025hypersigma} \href{https://github.com/WHU-Sigma/HyperSIGMA}{[code]} & SSL-M & TPAMI2025 & HSI & A FM for HSI with MIM pre-training on spatial and spectral dimensions.\\
       HyperFree\cite{li2025hyperfree} \href{https://github.com/Jingtao-Li-CVer/HyperFree}{[code]} & Sup. & CVPR2025 & HSI & A channel-adaptive, tuning-free FM for HSI enabling promotable and semantic-aware segmentation. \\
    \hline
    \end{tabular}
\end{table*}

\acrshort{vfms} are large-scale pre-trained FMs tailored for visual tasks for processing images and videos. In the RS domain, VFMs are expected to handle a diverse array of visual sensor modalities beyond RGB images, such as optical \acrshort{msi}, \acrshort{hsi}, \acrshort{sar} images, thermal images, and 3D \acrshort{lidar} point clouds. Recent advancements in VFMs for RS can be categorized into two main approaches: pre-training models (encompassing both supervised and self-supervised methods) and \acrfull{sam}-based models \cite{kirillov2023segment}.

\subsection{VFM Pre-training in RS}\label{sec.sup}

Pre-training has become a dominant approach in deep learning, where models are initially trained on a surrogate task before being fine-tuned for specific downstream applications \cite{bommasani2021opportunities}. This approach enhances knowledge transfer, enabling reduced computational costs and faster convergence during downstream task training. The success of FM pre-training relies on the availability of large-scale datasets and effective pre-training objectives. In the RS domain, advancements are accelerating, with notable progress in both \textit{datasets} and \textit{methods}.

\subsubsection{Pre-training Datasets}

Table \ref{tab. pre-train-datasets} provides an overview of the pre-training datasets utilized in recent \acrshort{rsfms}. A clear trend towards larger datasets is evident, with an increasing availability of training data and a broader inclusion of diverse RS modalities. It is promising to see these datasets being collected from various sources and featuring different \acrfull{gsd}, creating more comprehensive databases for RS research. On the other hand, the high cost of annotation poses practical limitations on the benefits of pre-training, which has driven a shift toward \acrfull{ssl} where pre-training datasets are unlabeled and easier to construct. Despite these advancements, RS pre-training datasets still fall short in scale—both in terms of size and modality diversity—when compared to those used in general foundation models.

Next, we review VFM pre-training methods in RS domains, including both supervised and self-supervised approaches. Table~\ref{tab:VFM-pretrain} summarizes representative methods.

\subsubsection{Supervised Pre-training}

Currently, most RS models are initialized using pre-trained parameters from ImageNet \cite{deng2009imagenet}, a computer vision dataset containing over 14 million \textit{natural} images across 1,000 categories. While this approach has led to significant results, the substantial domain gap between natural images and RS imagery—arising from differences in modality, perspective, color, texture, layout, etc.—often results in suboptimal pre-training performance for various RS tasks.

Some studies explored pre-trained models specifically tailored for RS \cite{wang2024mtp}.
For instance, Wang et al. \cite{wang2022empirical} collected the MillionAID dataset that contains millions of RS imagery scenes, and conducted an empirical examination on supervised pre-training on this dataset. They evaluated various deep architectures, including different CNNs and vision transformers, and found that RS-specific pre-training helps mitigate the data discrepancies encountered with ImageNet pre-training. However, they noted that some discrepancies may persist depending on the RS task, and their pre-training was limited to the classification of RGB images. More recently, Bastani et al. \cite{bastani2023satlaspretrain} investigated supervised multi-task pre-training using a unified model capable of learning from seven different label types, including classification, semantic segmentation, regression, object detection, instance segmentation, polyline prediction, and property prediction. This approach demonstrated clear performance improvements across various downstream RS tasks, even with differing image resolutions.

\subsubsection{Self-Supervised Pre-training}\label{sec.SSL}

\acrshort{ssl} follows a two-stage process:
(1) \textit{self-supervised pre-training}, where models learn useful and transferable representations from unlabeled data through various unsupervised pretext tasks, and (2) \textit{fine-tuning}, where these learned representations are adapted to specific downstream tasks, resulting in faster convergence and enhanced performance. Given the vast amount of RS and \acrshort{eo} data continuously captured by various platforms, labeling this data is highly resource-intensive, requiring substantial human effort, cost, and specialized expertise. This data-rich yet label-scarce environment positions SSL as an efficient and promising alternative for pre-training in RS, offering a cost-effective way to leverage the abundance of unlabeled data.

\begin{figure}[t]
    \centering
    \begin{subfigure}[b]{0.23\textwidth}
        \includegraphics[width=\textwidth]{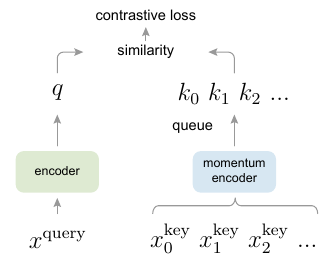}
        \caption{MoCo \cite{he2020momentum}}
    \end{subfigure}
    \hspace{1mm}
    \begin{subfigure}[b]{0.23\textwidth}
        \includegraphics[width=\textwidth]{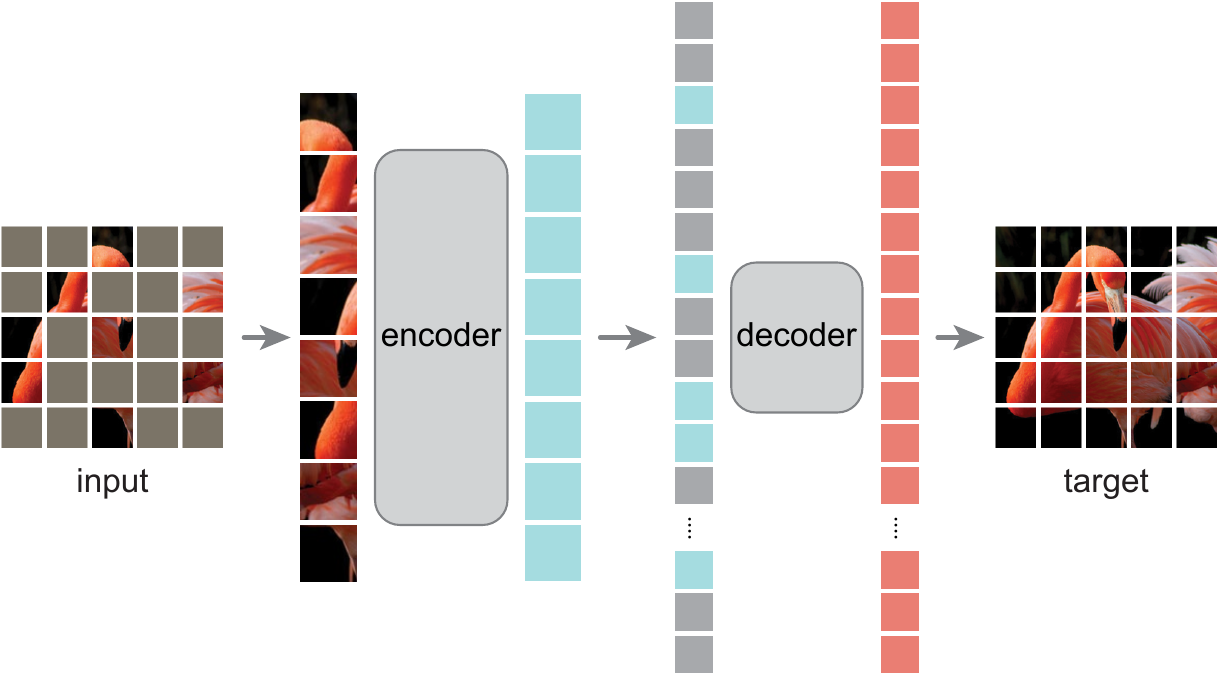}
        \caption{MAE \cite{he2022masked}}
    \end{subfigure}
    \caption{Typical approaches for two SSL methods.}
    \label{fig:ssl}
\end{figure}

Inspired by advancements in \acrshort{cv} \cite{jing2020self,xiao2023unsupervised}, SSL has made significant strides in RS, with two primary approaches leading the way: \textit{contrastive learning} and \textit{\acrfull{mim}}. While SSL has been widely explored in RS, much of this research was conducted in the pre-FM era. In contrast, this paper focuses specifically on SSL in the context of pre-training FMs. For a broader and more traditional survey of SSL techniques in RS, please refer to \cite{wang2022self}.

\noindent \textbf{Pre-training with Contrastive Objectives.} The core idea behind contrastive learning in SSL is to bring positive sample pairs (different augmented views of the same sample) closer together in the feature space while pushing negative pairs (different samples) apart, as illustrated in Fig.~\ref{fig:ssl}~(a). This approach effectively models data similarities and dissimilarities, resulting in representations that are robust and transferable. 

Given a batch of $B$ images, the contrastive learning objectives, such as InfoNCE \cite{oord2018representation} and its variants \cite{he2020momentum,chen2020simple}, are typically formulated as follows:
\begin{equation} 
\mathcal{L}^{\mathrm{InfoNCE}}_{I}=-\frac{1}{B}\sum^{B}_{i=1}\log\frac{\exp{(z_i^I \cdot z^I_{+}/\tau)}}{\sum^{B}_{j=1, j \neq i}\exp{(z_i^I \cdot z^I_j/\tau)}}, 
\end{equation}
where $z_i^I$ is the query embedding, $\{z_j^I\}^{B+1}_{j=1, j \neq i}$ are the key embeddings, with $z_+^I$ representing the positive key for $z_i^I$ and the remaining are treated as negative keys. The hyperparameter $\tau$ controls the smoothness or separation of the learned representations.
While contrastive pre-training \cite{chen2020simple,he2020momentum,chen2020improved,grill2020bootstrap,caron2020unsupervised} has made significant contributions in computer vision, it has also inspired numerous recent efforts in RS that can be grouped by sensor modalities: 

\textit{Optical RGB}: Majority of studies focus particularly on the optical RGB modality \cite{jean2019tile2vec}
For example, GASSL \cite{ayush2021geography} enhanced the MoCo-v2 framework \cite{chen2020improved} by integrating geo-location prediction as an additional pretext task. CACo \cite{mall2023change} utilized contrastive learning to detect both short-term and long-term changes, leveraging the spatiotemporal structure of temporal RS image sequences. MATTER \cite{akiva2022self} employed multi-temporal, spatially aligned RS imagery over unchanged regions to develop representations invariant to illumination and viewing angles, specifically for materials and textures. CSP \cite{mai2023csp} introduced contrastive spatial pre-training for geo-tagged images, enriching representation learning through geo-location information. Furthermore, Huang et al. \cite{huang2024generic} combined self-supervised contrastive learning on unlabeled RS images with supervised learning on labeled natural images, demonstrating that integrating generic knowledge can significantly enhance pre-training for RS imagery.

\textit{Non-optical RGB}: Several studies have expanded contrastive learning to include modalities beyond optical RGB. For instance, SeCo \cite{manas2021seasonal} applied contrastive learning to \acrshort{msi}, enabling the learning of time- and position-invariant representations over seasonal changes. Li et al. \cite{li2021geographical} incorporated global land cover products and the geographical locations of RS images to provide additional supervision in their contrastive learning framework. Additionally, Li et al. \cite{li2022global} implemented both global and local contrastive learning techniques on RGB and \acrshort{nir} images. SkySense \cite{guo2024skysense} introduced a contrastive learning framework that integrates multiple modalities and spatial granularities, leveraging geo-context prototype learning to facilitate cross-modal processing among RGB, MSI, and \acrshort{sar} images.

\noindent \textbf{Pre-training with Generative Objectives.} Generative pre-training aims to develop rich, general representations for downstream tuning using unlabeled data through generative tasks. A widely used method is Masked Image Modeling (MIM), depicted in Fig.~\ref{fig:ssl}~(b), where random patches of an input image are masked, and the model is tasked with reconstructing the missing areas in pixel space \cite{he2022masked}. This reconstruction process requires the model to understand the objects and their surrounding context within the image, resulting in feature representations that are both effective and advantageous for downstream applications. The loss function for a batch of $B$ images is defined as:
\begin{equation}
    \mathcal{L}_{MIM}=-\frac{1}{B}\sum_{i=1}^{B}\log f_\theta(\overline{x}^I_i|\hat{x}^I_i)
\end{equation}
where $\overline{x}^I_i$ and $\hat{x}^I_i$ represent the masked and unmasked patches in $x^I_i$, respectively, and $f_\theta$ refers to the mean squared error (MSE) between the reconstructed and original images, calculated only on the masked patches \cite{he2022masked}.

MIM has been widely applied in the RS domain; however, the original MAE framework \cite{he2022masked}, which is designed for static natural RGB images, often struggles with the unique characteristics of RS imagery. To this end, a series of studies have been proposed to address these RS-specific challenges based on the MIM architectures:
\begin{itemize}[noitemsep,nolistsep,topsep=0pt,parsep=0pt,partopsep=0pt]
    \item \textit{Varying temporal resolutions}: SatMAE \cite{cong2022satmae} focused on MAE pre-training with temporal embedding along with independently masking image patches across time.
    \item \textit{Varying spatial resolutions}: SatMAE++ \cite{noman2024rethinking} proposed multi-scale reconstructions for both RGB and MSI for addressing spatial resolution variability in RS images. Some approaches also explore reconstruction in the \textit{frequency} domain \cite{tang2023cross,dong2024generative}. For instance, Scale-MAE \cite{tang2023cross} combines low/high frequency reconstruction with ground sample distance positional encoding.
    \item \textit{Varying object sizes and  orientations}: Sun et al. \cite{sun2022ringmo} introduced a patch-incomplete masking strategy to tackle small object sizes common in RS images, which random masking can overlook. Wang et al. \cite{wang2022advancing} developed a transformer with rotated, varied-size window attention, paired with MAE pre-training. Alternatively, MA3E \cite{li2024masked} pre-trains models using angle-aware MIM to learn angle-invariant representations.
    \item \textit{Varying spectral dimensions}: researches have been explored in masking along the \textit{spectral} dimension for MSI \cite{hong2024spectralgpt}, HSI \cite{wang2025hypersigma}, and multimodal data \cite{xiong2024neural}.
\end{itemize}

Additionally, efforts have been made to develop \textit{efficient} MAE frameworks tailored to the RS domain: Mendieta et al. \cite{mendieta2023towards} investigated continual pre-training using ImageNet models as an auxiliary distillation objective to enhance geospatial FMs while reducing downstream computational costs; Wang et al. \cite{wang2024scaling} introduced an efficient MAE that encodes and reconstructs only a subset of semantically rich patch tokens.

There is also research exploring multimodal MAE for paired RS data aligned with geographic coordinates. For example, msGFM \cite{han2024bridging} designed a multimodal MIM framework with a shared encoder and distinct patch embedding/decoder, capable of processing RGB, MSI, SAR, and \acrshort{dsm} data. MMEarth \cite{nedungadi2024mmearth}, on the other hand, developed a multimodal MAE capable of handling 12 paired pixel-level and image-level modalities.

\noindent\textbf{Pre-training with Hybrid Objectives.} Several studies have integrated contrastive learning with MIM to leverage both discriminative and generative representations. For example, Muhtar et al. \cite{muhtar2023cmid} proposed a contrastive masked image distillation method, which pre-train models by combining contrastive learning with MIM in a self-distillation framework within the RGB space. CROMA \cite{fuller2024croma} separately encodes masked spatial- and temporal-aligned MSI and SAR data, then employs cross-modal contrastive learning with an additional encoder that fuses the sensor data to generate joint multimodal encodings. These encodings are subsequently used to reconstruct the masked patches via a lightweight decoder. Cross-Scale MAE \cite{tang2023cross} employs scale augmentation and enforces cross-scale consistency through a combination of contrastive and generative losses. These approaches illustrate that merging contrastive and generative SSL enables the learning of complementary representations, thereby enhancing the effectiveness of downstream RS tasks.

\subsection{SAM for RS}

\subsubsection{SAM Overview}
\begin{figure}[t]
    \centering
    \includegraphics[width=\linewidth]{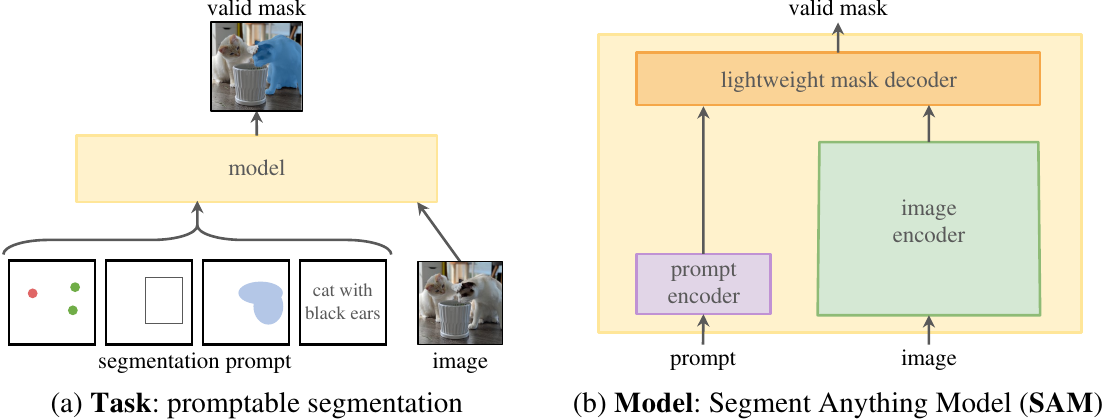}
    \caption{Overview of SAM \cite{kirillov2023segment}, a foundation model for general mask segmentation: (a) promptable segmentation \textit{task} framework, and (b) SAM's \textit{model} architecture. The figure is adapted from \cite{kirillov2023segment}.}
    \label{fig:sam}
\end{figure}

\acrfull{sam}~\cite{kirillov2023segment} represents a significant advancement in image segmentation. Trained on the extensive SA-1B dataset, which includes over 11 million RGB images and 1.1 billion mask annotations, SAM excels in promptable segmentation. As shown in Fig.~\ref{fig:sam} (a), it allows users to generate high-quality object segmentation with reference of various geometric prompts, such as points, bounding boxes, or coarse masks.

SAM has significantly impacted computer vision, particularly image segmentation, due to its strong generalization capabilities, robust zero-shot performance, and prompt-based flexibility. Its versatility makes it adaptable to a wide array of downstream applications such as medical checkups and autonomous vehicles, aligning with the trend toward AI models that can handle multiple tasks seamlessly.

As illustrated in Fig.~\ref{fig:sam}~(b), SAM consists of three core components: (1) a heavyweight \textit{image encoder} (based on ViT\cite{dosovitskiy2020image}) that transforms input images into embeddings, (2) a lightweight \textit{prompt encoder} that processes geometric prompts into prompt embeddings, and (3) a lightweight \textit{mask decoder} that integrates these embeddings to generate accurate segmentation masks. A more detailed description can be found in \cite{kirillov2023segment}.

\subsubsection{SAM-based studies for RS}
SAM's powerful segmentation capabilities have garnered significant attention within the RS community. However, its performance in RS is limited compared to natural images due to several key challenges:
\begin{itemize}[noitemsep,nolistsep,topsep=0pt,parsep=0pt,partopsep=0pt]
    \item RS images, typically captured from aerial or satellite views from large scale areas, encompass vast and intricate \acrshort{lulc} patterns that differ substantially from SAM’s training data in natural domains.
    \item  SAM is optimized for general mask segmentation without semantic understanding, which is critical for interpreting geographic and environmental data in RS applications.
    \item  SAM is primarily designed for RGB images, whereas RS data often includes non-RGB sensor modalities like \acrshort{msi}, \acrshort{hsi}, or \acrshort{sar}.
\end{itemize}

To address these challenges, several adaptations of SAM have been proposed. Some focus on extending SAM for \textit{semantic segmentation} ability in RS. For example, Wang et al. \cite{wang2023samrs} developed a pipeline combining SAM with common RS object detection datasets to generate semantic segmentation datasets for optical RGB images. 
Ma et al. \cite{ma2024sam} used SAM’s object and boundary predictions to regularize the outputs of other semantic segmentation models trained on labeled RS images. 
UV-SAM \cite{zhang2024uv} refined coarse masks generated through class activation mapping into accurate pseudo-labels, which were then used to train a semantic segmenter for urban village segmentation. 

Beyond semantic segmentation, SAM has also been applied to \textit{change detection} tasks \cite{wang2023cs, ding2024adapting,zheng2024segment}. For example, Zheng et al. \cite{zheng2024segment} explore zero-shot change detection built upon SAM;
Chen et al. \cite{chen2024rsprompter} introduced a learnable anchor-based prompter and mask decoder, enabling SAM to process categorical inputs for \textit{instance segmentation}. 

Unlike previous studies focusing on RGB spaces, some research has explored extending semantic segmentation to the \textit{non-RGB} domain. For example, Yan et al. \cite{yan2023ringmo} adapted SAM for multiple RS modalities (RGB, SAR, PolSAR, DSM, MSI) by freezing SAM’s encoders, applying LoRA for \acrshort{peft}, and integrating separate decoders for each modality, achieving semantic segmentation across diverse sensor types. Similarly, Osco et al. \cite{osco2023segment} combined SAM with GroundDINO \cite{liu2023grounding}, an open-set object detection model, to perform text-based semantic segmentation on unmanned aerial vehicles, airborne, and satellite RGB images.

While most of these approaches rely on fully supervised methods, requiring substantial labeled data, CAT-SAM \cite{xiao2024catsam} provides a data-efficient alternative for \textit{few-shot adaptation}. 
It employs a prompt bridge structure that jointly fine-tunes SAM’s image encoder and mask decoder through a conditional joint tuning design. This method resolves tuning imbalances between the encoder and decoder, achieving superior segmentation across multiple RS sensor modalities even with very limited downstream tuning samples.

% Furthermore, SAM’s design for single-modal RGB images limits its application to the multimodal data often used in RS, which typically incorporates data from multiple sensor types. 

\subsection{Summary and Discussion}

In summary, research on VFMs in RS predominantly focuses on pre-training strategies, both supervised and unsupervised, with a strong emphasis on SSL. SSL approaches, such as contrastive learning and generative MIM, have been key in reducing reliance on expensive and time-consuming annotations. However, the development of widely adopted pre-trained models is constrained by the limited scale and diversity of current SSL datasets in RS, whether for individual modalities or multimodal applications. 

Additionally, the introduction of SAM has spurred numerous adaptations tailored to RS data, extending its utility across different RS-specific scenarios, semantic recognition tasks, cross-modal transfer, and multimodal processing.
While there are a few studies exploring other VFMs, such as \textit{depth anything model} \cite{yang2024depth} for tasks like canopy height estimation \cite{cambrin2024depth} and building extraction \cite{chen2024depth}, progress in these areas has been gradual. 
% Significant advancements are still anticipated to unlock the full potential of VFMs in RS.

It is noted that while many of reviewed frameworks have been trained and tested in multiple sensor modalities individually, only a few RSFMs have been developed to accept multiple sensor modalities simultaneously as complementary input. For instance, msGFM \cite{han2024bridging} demonstrates that multimodal pre-training can enhance downstream performance in multimodal inference tasks following fine-tuning. Another example is TerraMind \cite{jakubik2025terramind}, which incorporates diverse modalities for paired data, including optical RGB, SAR, NDVI, \acrshort{lulc}, \acrshort{dsm}, text captions, and geolocation coordinates. Leveraging a globally available public dataset, TerraMind conducts large-scale multimodal MIM pre-training and achieves promising results in cross-modal synthetic data generation, as well as in zero-shot and few-shot scenarios. MM-SAM \cite{xiao2024segment} expands SAM’s RGB processing capabilities to support cross-modal and multimodal processing, enabling enhanced segmentation across a variety of sensor suites commonly used in RS, such as RGB plus \acrshort{hsi}, RGB plus \acrshort{lidar}, etc.
Nevertheless, the performance gains reported in these studies over single-modality fine-tuning remain relatively modest. This suggests that, while multimodal learning is a promising direction, the effective integration of heterogeneous sensor modalities continues to pose significant challenges, largely due to the intrinsic differences in spatial resolution, spectral characteristics, and physical sensing principles across modalities.

Importantly, VFMs have demonstrated new capabilities that surpass traditional RS approaches. Examples include any-to-any generative frameworks across multiple RS sensor modalities and products \cite{jakubik2025terramind} via multimodal MIM, accelerated semantic mask annotation pipelines based on SAM \cite{wang2023samrs}, and achieving high segmentation performance with minimal fine-tuning data \cite{xiao2024catsam}. These advancements suggest that VFMs are beginning to accelerate the evolution of RS interpretation tasks. Although current RSFM studies have not yet fundamentally reshaped a specific RS application that traditionally addressed by conventional approaches, comparable to the paradigm shifts brought by GPT-4 or SAM in the general domain, the future holds substantial promise for unlocking the full potential of VFMs in RS.

\section{Vision-Language Models for RS}\label{sec:vlm}

\begin{figure}[t]
    \centering
    \includegraphics[width=\linewidth]{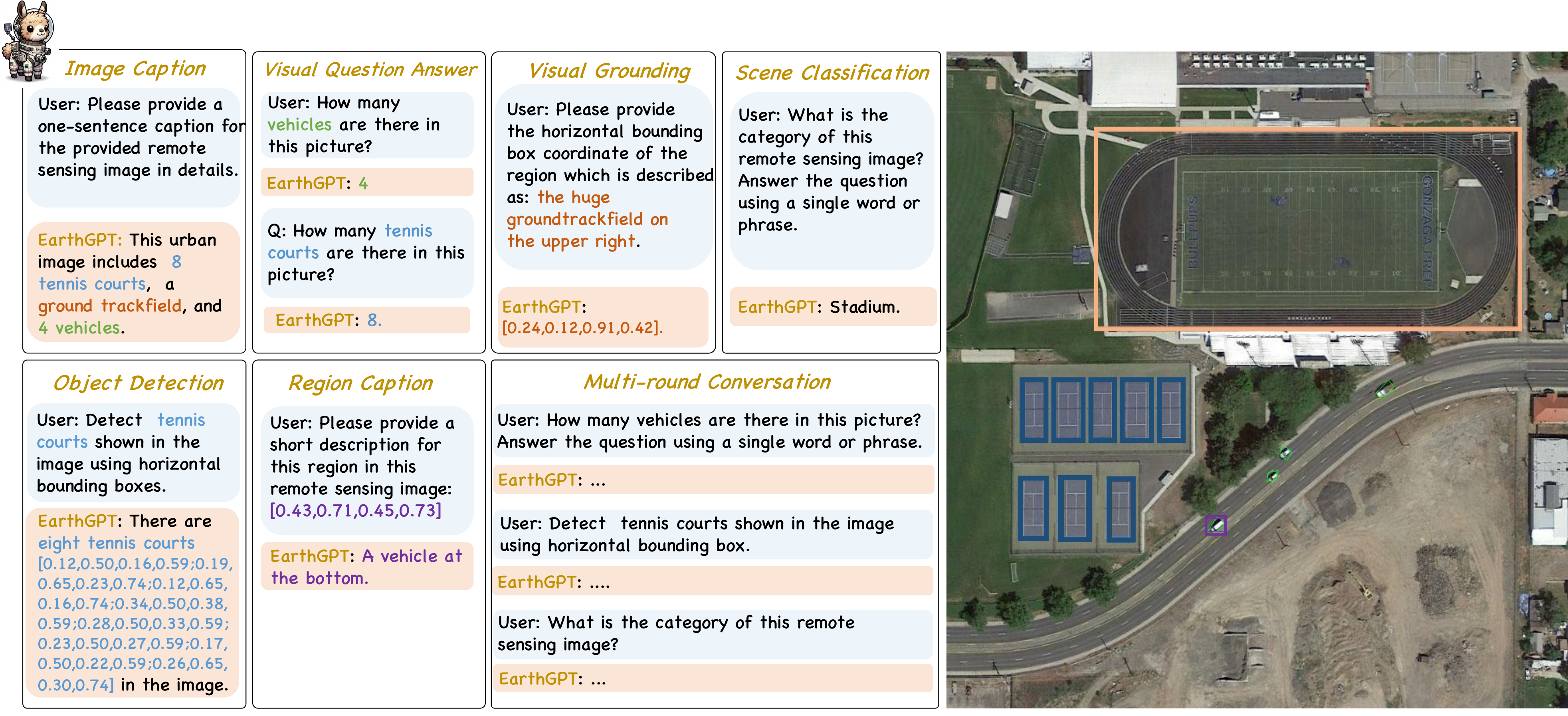}
    \caption{Multi-task outputs of the generative VLM. The figure is sourced from \cite{zhang2024earthgpt}.}
    \label{fig:vlm-generative-result}
\end{figure}

\acrfull{vlms} are multimodal models designed to integrate and process both visual and textual data. In contrast to \acrshort{vfms}, which focus exclusively on visual data, VLMs incorporate language understanding, unlocking powerful semantic interpretation and human-system interaction facilitated by textual representations. By training on extensive image-text datasets, VLMs excel in tasks requiring deep comprehension of both visual content and linguistic context, such as image classification~\cite{radford2021learning, openclip, evaclip}, segmentation~\cite{li2022language, zhou2023zegclip}, visual grounding~\cite{zhang2023llavagrounding, bai2023qwen, Qwen2-VL}, and image captioning~\cite{wang2022git, li2022blip, li2023blip2}, bridging the gap between visual perception and language-based interpretation.
Fig.~\ref{fig:vlm-generative-result} gives examples of VLM tasks in RS domains.

\subsection{Preliminaries of VLMs in Natural Domains}

\begin{table*}[t]
\caption{Summary of remote sensing vision-language generative datasets} 
    \label{tab:VLM-RSDataset}
    \setlength\tabcolsep{1pt}
    \begin{adjustbox}{max width=\textwidth}
    \begin{tabular}{|l|l|c|c|c|c|l|c|}
        \hline
        Task & Dataset & Image Size & GSD (m) & \#Text & \#Images & Content & Link \\ 
        \hline
        \multirow{10}{*}{VQA} & RSVQA-LR \cite{lobry2020rsvqa} & 256 & 10 & 77K & 772 & Questions for existing judging, area estimation, object comparison, scene recognition &  \href{https://zenodo.org/records/6344334}{\faDownload} \\
        & RSVQA-HR  \cite{lobry2020rsvqa} & 512 & 0.15 & 955K & 10,659 & Questions for existing judging, area estimation, object comparison, scene recognition &  \href{https://zenodo.org/records/6344367}{\faDownload} \\
        & RSVQAxBen\cite{lobry2021rsvqa} & 120 & 10--60 & 15M & 590,326 & Questions for existing judging, object comparison, scene recognition & \href{https://zenodo.org/records/5084904}{\faDownload} \\
        & RSIVQA\cite{zheng2021mutual} & 512--4,000 & 0.3--8 & 111K & 37,000 & Questions for existing judging, area estimation, object comparison, scene recognition & \href{https://github.com/spectralpublic/RSIVQA}{\faDownload} \\
        & HRVQA\cite{li2024hrvqa} & 1,024 & 0.08 & 1,070K & 53,512 & Questions for existing judging, object comparison, scene recognition &   \href{https://hrvqa.nl}{\faDownload} \\
        & CDVQA\cite{yuan2022change} & 512 & 0.5--3 & 122K & 2,968 & Questions for object changes & \href{https://github.com/YZHJessica/CDVQA}{\faDownload} \\
        & FloodNet\cite{rahnemoonfar2021floodnet} & 3,000--4,000 & - & 11K & 2,343 & Questions for for building and road damage assessment in disaster scenes &  \href{https://github.com/BinaLab/FloodNet-Challenge-EARTHVISION2021}{\faDownload} \\
        & RescueNet-VQA\cite{sarkar2023rescuenet} & 3,000--4,000 & 0.15 & 103K & 4,375 & Questions for building and road damage assessment in disaster scenes  & \href{https://www.codabench.org/competitions/1550}{\faDownload} \\
        & EarthVQA\cite{wang2024earthvqa} & 1,024 & 0.3 & 208K & 6,000 & Questions for relational judging, relational counting, situation analysis, and comprehensive analysis & \href{https://github.com/Junjue-Wang/EarthVQA}{\faDownload} \\
        \hline
        % \multirow{4}{*}{Image-Text Pre-tranining} & 
        Image-Text & RemoteCLIP~\cite{liu2024remoteclip} & varied & varied & not specified & not specified & Developed based on retrieval, detection and segmentation data & \href{https://github.com/ChenDelong1999/RemoteCLIP}{\faDownload} \\
        % & GRAFT NAIP~\cite{mall2023remote} & 224 & 1 &10.2M & 0 & Ground-satellite image pairs, no text\\
        % & GRAFT Sentinel-2~\cite{mall2023remote} & 224 & 10 & 8.7M & 0 & Ground-satellite image pairs, no text\\
        Pre-tranining & RS5M~\cite{zhang2023rs5m} & not specified & varied & 5M & 5M & Filtered public datasets, captioned existing data & \href{https://github.com/om-ai-lab/RS5M}{\faDownload}\\
        & SKyScript~\cite{wang2024skyscript} & not specified & 0.1 - 30 & 2.6M & 2.6M & Earth Engine images linked with OpenStreetMap semantics & \href{https://github.com/wangzhecheng/SkyScript}{\faDownload}\\  
        \hline
        \multirow{8}{*}{Captioning} & RSICD\cite{lu2017exploring} & 224 & - & 24,333 & 10,921 & Urban scenes for object description & \href{https://github.com/201528014227051/RSICD_optimal}{\faDownload} \\
        & UCM-Caption\cite{qu2016deep} & 256 & 0.3 & 2,100 & 10,500 & Urban scenes for object description & \href{https://pan.baidu.com/s/1mjPToHq#list/path=\%2F}{\faDownload} \\
        & Sydney\cite{qu2016deep} & 500 & 0.5 & 613 & 3,065 & Urban scenes for object description & \href{https://pan.baidu.com/s/1hujEmcG#list/path=\%2F}{\faDownload} \\
        & NWPU-Caption\cite{cheng2022nwpu}  & 256 & 0.2-30 & 157,500 & 31,500 & Urban scenes for object description & \href{https://github.com/HaiyanHuang98/NWPU-Captions}{\faDownload} \\ 
        & RSITMD\cite{yuan2021exploring}  & 224 & - & 4,743 & 4,743 & Urban scenes for object description & \href{https://github.com/xiaoyuan1996/AMFMN/tree/master/RSITMD}{\faDownload} \\
        & RSICap\cite{hu2023rsgpt} & 512 & varied & 3,100  & 2,585 & Urban scenes for object description &  \href{https://github.com/Lavender105/RSGPT}{\faDownload} \\
        & ChatEarthNet\cite{zhan2023rsvg} & 256 & 10 & 173,488  & 163,488 & Urban and rural scenes for object description &  \href{https://github.com/zhu-xlab/ChatEarthNet}{\faDownload} \\ 
        \hline
        Visual & GeoVG\cite{sun2022visual} & 1,024 & 0.24--4.8   & 7,933 & 4,239 & Visual grounding based on object properties and relations &  \href{https://drive.google.com/file/d/1kgnmVC6FVKdxCwaoG77sOfkaIHS_XiFt/view}{\faDownload} \\ 
        Grounding & DIOR-RSVG\cite{zhan2023rsvg} & 800 & 0.5--30 & 38,320  & 17,402 & Visual grounding based on object properties and relations &  \href{https://drive.google.com/drive/folders/1hTqtYsC6B-m4ED2ewx5oKuYZV13EoJp_}{\faDownload} \\ 
        \hline
        \multirow{2}{*}{Mixed } & MMRS-1M\cite{sun2022visual} & varied & varied   & 1M & 975,022 & Collections of RSICD, UCM-Captions, FloodNet, RSIVQA, UC Merced, DOTA, DIOR-RSVG, etc  & \href{https://github.com/wivizhang/EarthGPT}{\faDownload} \\
        & Geochat-Set\cite{zhan2023rsvg} & varied & varied & 318k & 141,246 & Developed based on DOTA, DIOR, FAIR1M, FloodNet, RSVQA and NWPU-RESISC45 &  \href{https://github.com/mbzuai-oryx/GeoChat}{\faDownload} \\
        Multi-task & LHRS-Align\cite{muhtar2024lhrs} & 256 & 1.0 & 1.15M & 1.15M & Constructed from Google Map and OSM properties & \href{https://github.com/NJU-LHRS/LHRS-Bot}{\faDownload} \\
        & VRSBench\cite{muhtar2024lhrs} & 512 & varied & 205,307 & 29,614 & Developed based on DOTA-v2 and DIOR dataset & \href{https://github.com/lx709/VRSBench}{\faDownload} \\
        & TEOChatlas\cite{irvin2024teochat} & varied & varied & 554K & 1,244,393 & Temporal EO dataset for multimodal instruction-tuning & \href{https://huggingface.co/datasets/jirvin16/TEOChatlas}{\faDownload}\\
        & SARLANG-1M~\cite{wei2025sarlang} & varied & 0.1-25 & 1,126,277 & 118,331 & High quality SAR image-text dataset for image captioning and VQA. & \href{https://huggingface.co/datasets/YiminJimmy/SARLANG-1M}{\faDownload}\\
        \hline
        \end{tabular}
        \end{adjustbox}
\end{table*}

\acrshort{vlms} typically consist of two core components: a \textit{visual encoder} and a \textit{language encoder}. The visual encoder converts raw images into dense feature representations, while the language encoder processes textual inputs into corresponding embeddings. These embeddings are then fused or aligned within a shared latent space, allowing the model to establish associations between visual and linguistic concepts. A crucial aspect of pre-training VLMs is bridging vision and language through image-text pairs, typically achieved via two main objectives: \textit{contrastive learning} and \textit{generative modeling} \cite{zhang2024vision}.

\noindent\textbf{Contrastive objectives} focus on aligning matched image-text pairs by bringing them closer together in the representation space while pushing mismatched pairs apart. A representative work CLIP~\cite{radford2021learning}, jointly trains an image encoder and a text encoder using contrastive learning. This approach enables CLIP to learn a unified representation space for both visual and textual data, resulting in exceptional zero-shot classification capabilities. CLIP's flexibility has made it highly effective in various open-vocabulary perception tasks~\cite{zhong2022regionclip, liang2023open, cheng2024yolo}. However, the original CLIP model was not fully open-sourced, prompting the development of OpenCLIP~\cite{openclip}, a fully open-source implementation. Building on these advances, EVA-CLIP~\cite{evaclip} introduces enhanced training strategies to improve performance and efficiency. These modifications allow EVA-CLIP to achieve impressive results over previous CLIP models while reducing computational overhead and enhancing training stability.

\noindent \textbf{Generative objectives} aim to train the model to generate coherent and relevant texts or images. Generally, there are two main approaches to these objectives: masked reconstruction and autoregressive next-token prediction.

\begin{itemize}[noitemsep,nolistsep,topsep=0pt,parsep=0pt,partopsep=0pt]
    \item \textit{Masked reconstruction}, used in models like FLAVA~\cite{singh2022flava} and MaskVLM~\cite{kwonmasked}, involves predicting masked tokens in text or patches in images. This technique improves the model’s ability to understand context and relationships across visual and linguistic modalities, thereby enhancing cross-modal comprehension.
    \item \textit{Autoregressive next-token prediction} trains the model to generate the next token in a sequence based on prior context. This has become a dominant paradigm in VLM training, largely due to its synergy with pre-trained \acrshort{llms}. VLMs benefit from the vast knowledge and language comprehension of LLMs while extending these capabilities to the visual domain. Exemplar models like LLaVA~\cite{liu2023llava} excel across various vision-language tasks. These models typically consist of three core components: a pre-trained language module (e.g., Llama2~\cite{touvron2023llama}, Vicuna~\cite{chiang2023vicuna}, Llama3~\cite{dubey2024llama}), a pre-trained visual encoder (e.g., CLIP~\cite{radford2021learning}, EVA-CLIP~\cite{evaclip}, SigLIP~\cite{siglip}), and a trainable connection module that bridges visual and language embeddings (e.g., Linear Projection~\cite{liu2023llava}, Two-layer MLP~\cite{liu2024llavanext}, Q-Former~\cite{li2023blip2}). This architecture allows the model to leverage the strengths of pre-trained visual and language modules, while the connection module learns to align information across these modalities.
    \item Some VLMs~\cite{li2022blip,li2023blip2,instructblip} adopt a \textit{hybrid} approach by combining multiple training objectives, leveraging the strengths of different methodologies to improve model robustness and adaptability across diverse vision-language tasks.
\end{itemize}

It is noted that recent efforts have extended this framework beyond image-text pairs, exploring combinations of LLMs with encoders for other modalities such as audio~\cite{chu2023qwenaudio, panagopoulou2023xinstructblip} and point clouds~\cite{xu2023pointllm, panagopoulou2023xinstructblip}. These advancements aim to create more comprehensive multimodal models that understand and generate textual content from a wider range of sensory inputs, shedding light on advanced multimodal understanding within RS domains.

\subsection{Vision-language datasets for RS}

Building on the advancements of VLMs in general domains, the RS field is actively developing RS-specific vision-language datasets as a foundation for technical progress. It's noted that vision-language tasks in RS have a long history, even prior the  era of FMs. To offer a comprehensive overview of these efforts, we summarize the existing vision-language datasets for RS in Table~\ref{tab:VLM-RSDataset}.

\noindent\textbf{RS VQA}: RS VQA datasets are designed to interpret user questions, identify relevant visual evidence, analyze spatial relationships among geospatial objects, and generate concise textual responses. In recent years, larger and more diverse datasets have been developed, particularly for applications like urban planning and disaster assessment. Notable examples include EarthVQA~\cite{wang2024earthvqa}, RSVQA~\cite{lobry2020rsvqa}, RSIVQA~\cite{zheng2021mutual}, and FloodNet~\cite{rahnemoonfar2021floodnet}.

\noindent\textbf{Image-Text Pre-training in RS}: Contrastive pre-training techniques like CLIP~\cite{radford2021learning}, have greatly advanced the creation of datasets that pair RS images with textual descriptions, enabling tasks like zero-shot classification and cross-modal retrieval. Several key datasets have emerged, each with a unique approach to generating large-scale image-text pairs for RS applications. For instance, RemoteCLIP~\cite{liu2024remoteclip} transforms annotations from detection and segmentation datasets into image-caption pairs. GRAFT~\cite{mall2023remote} links RS imagery from NAIP and Sentinel-2 with co-located ground-level images collected from the internet. RS5M~\cite{zhang2023rs5m} filters and captions existing RS datasets to enrich data for visual-language tasks, while SkyScript~\cite{wang2024skyscript} connects Google Earth Engine images with OpenStreetMap data using geographic coordinates. These datasets demonstrate a concerted effort to repurpose and integrate existing resources for emerging vision-language tasks, significantly expanding the range of RS applications.

\noindent \textbf{RS Image Captioning:} RS image captioning generates descriptive sentences of variable lengths to summarize the objects and features within an RS image, without needing specific instructions or questions as input. Many of the available RS captioning datasets are secondary developed from scene classification datasets, such as NWUPU45-Caption~\cite{cheng2022nwpu} from NWUPU45 dataset~\cite{cheng2017remote} and UCM-Caption~\cite{qu2016deep} from UCM dataset\cite{yang2010bag}.

\noindent\textbf{RS Visual Grounding:} RS visual grounding involves identifying the correct geospatial object in an image based on a given textual description. The model must accurately locate the target object while filtering out irrelevant ones that do not meet the specified criteria. Datasets for RS visual grounding include GeoVG\cite{sun2022visual} and DIOR-RSVG\cite{zhan2023rsvg} datasets.

It is encouraging to see rapid progress in developing visual-language datasets for RS. However, the field still lags behind general domains and falls short of meeting the high demands of \acrshort{eo} applications. Additionally, most efforts remain concentrated on RGB imagery, with limited exploration of other important RS modalities such as \acrshort{sar}, \acrshort{msi}, and \acrshort{hsi} data.

\subsection{VLMs with Contrastive Objectives}

Similar to the contrastive learning in VFMs discussed in Section \ref{sec.SSL}, the contrastive pre-training in VLMs~\cite{radford2021learning} also employs the InfoNCE loss~\cite{oord2018representation} as the training objective. 
A typical example of FM is CLIP~\cite{radford2021learning}.
The primary difference is that, in CLIP, the contrastive samples are image-text pairs, and the objective becomes a symmetric image-text InfoNCE loss, defined as:
\begin{equation}
    \mathcal{L}^{IT}_{infoNCE} =\mathcal{L}_{I\rightarrow T} + \mathcal{L}_{T\rightarrow I},
\end{equation}
Here, the first term $L_{I\rightarrow T}$ contrasts the query image with text keys (i.e., \textit{image-to-text}), while the second term $L_{T\rightarrow I}$ contrasts the query text with image keys (i.e., \textit{text-to-image}). 
Given a batch of $B$ image-text pairs, $L_{I\rightarrow T}$ and $L_{T\rightarrow I}$ are defined as:
\begin{equation} 
\mathcal{L}_{I\rightarrow T}=-\frac{1}{B}\sum^{B}_{i=1}\log\frac{\exp{(z_i^I \cdot z_i^T/\tau)}}{\sum^{B}_{j=1}\exp{(z_i^I \cdot z_j^T/\tau)}}, 
\end{equation}
and 
\begin{equation} 
\mathcal{L}_{T\rightarrow I}=-\frac{1}{B}\sum^{B}_{i=1}\log\frac{\exp{(z_i^I \cdot z_i^T/\tau)}}{\sum^{B}_{j=1}\exp{(z_i^T \cdot z_j^I/\tau)}}, 
\end{equation}
where $z_I$ and $z_T$ are the image and text embeddings, respectively, and $\tau$ is a learnable temperature parameter. Minimizing this loss aligns image and text representations, ensuring that matching pairs have higher similarity scores than mismatched pairs. This process enables CLIP to learn rich vision-language representations that can be effectively applied to various tasks without requiring task-specific fine-tuning.

Although CLIP demonstrates impressive performance across a wide range of visual tasks, its training data primarily comprises general-purpose image-text pairs sourced from the internet. While this broad approach is powerful, it may not fully capture the unique attributes and characteristics of RS data. To address this limitation, researchers have developed adaptations of CLIP specifically tailored for RS applications. These adaptations generally fall into two categories: \textit{visual-text models} and \textit{visual-location models}, which will be reviewed in the remaining of this subsection.

\subsubsection{Visual-Text Models}

Existing studies in this direction primarily extend CLIP into RS domains, with the focus largely on developing RS-specific datasets and benchmarks. These efforts enable the fine-tuning of CLIP/OpenCLIP, either through fully-supervised approaches or more efficient, resource-conscious fine-tuning strategies tailored to RS tasks.

For instance, RemoteCLIP~\cite{liu2024remoteclip} adapts CLIP via \textit{continual pretraining} while preserving its original architecture. Evaluated on tasks such as classification and cross-modal retrieval, it also introduces the RemoteCount benchmark for object counting. Similarly, SkyCLIP~\cite{wang2024skyscript} also utilizes continual pretraining on its custom dataset, assessing the model's performance on zero-shot scene classification, fine-grained attribute classification, and cross-modal retrieval. Additionally, GeoRSCLIP~\cite{zhang2023rs5m} explores both full and \acrshort{peft} techniques, targeting zero-shot classification, cross-modal retrieval, and semantic localization across multiple benchmarks.

Differently, GRAFT~\cite{mall2023remote} introduces a unique strategy without textual annotations. By aligning satellite images with ground-level imagery using contrastive learning, GRAFT builds both image-level and pixel-level vision-language models for RS. Evaluated across classification, retrieval, and segmentation tasks, it shows considerable improvements, especially in zero-shot settings, leveraging the alignment between satellite and ground-level images.

\subsubsection{Visual-Location Models}
Another important research direction in contrastive-based VLMs for RS involves modeling RS data with geographical location information like latitude and longitude. These visual-location models aim to integrate RS imagery with geometric location context, offering a more interactive and comprehensive understanding of \acrshort{eo} data.

CSP~\cite{mai2023csp} is a pioneering study, using a dual-encoder architecture to process geo-tagged images and locations separately, then aligning their embeddings through contrastive learning in a self-supervised manner. CSP explores different sampling strategies for positive and negative pairs and evaluates various self-supervised losses, demonstrating strong performance in geo-aware classification especially in few-shot learning setups over iNat2018~\cite{van2018inaturalist} and fMoW~\cite{christie2018functional} datasets, without relying on textual annotations.

GeoCLIP~\cite{vivanco2024geoclip} takes a different approach by framing geo-localization as image-to-GPS retrieval. It incorporates an encoder based on equal earth projection, random Fourier features, and hierarchical representation, leading it to represent GPS coordinates as a continuous function and bypass the limitations of predefined geographic classes. By combining this with a CLIP-based image encoder, GeoCLIP achieves precise GPS retrieval rather than simple region classification, offering flexible and accurate localization, even in low-data scenarios, with potential applications beyond geo-localization.

SatCLIP~\cite{klemmer2023satclip} emphasizes geographic diversity in its sampling. Unlike CSP, it uses the globally sampled S2-100K dataset, which comprises 100,000 Sentinel-2 satellite images to ensure broad geographic coverage. The model utilizes a spherical harmonics-based location encoder to align satellite images with geographic embeddings, enhancing generalization across diverse regions. This method demonstrates solid performance in various location-dependent tasks, such as temperature prediction and population density estimation, contributing to the development of geographically balanced models capable of generalizing globally.

SatMIP\cite{bourcier2024learning} encodes metadata like location and time as textual captions and aligns these with images in a shared embedding space through a metadata-image contrastive learning task. This approach enables the model to learn robust image representations, improving its ability to handle recognition tasks by leveraging the rich contextual information provided by metadata.

\subsection{VLMs with Generative Objectives}

Generative objectives in VLMs allow networks to learn semantic features by generating data, whether images, text, or cross-modal outputs \cite{zhang2024vision}. While the RS field has been slower to adopt these techniques, current studies on VLMs with generative objectives in RS largely build upon the LLaVA~\cite{liu2023llava} framework, a notable open-source VLM designed for image-conditioned text generation.

\begin{figure}[t]
    \centering
    \includegraphics[width=\linewidth]{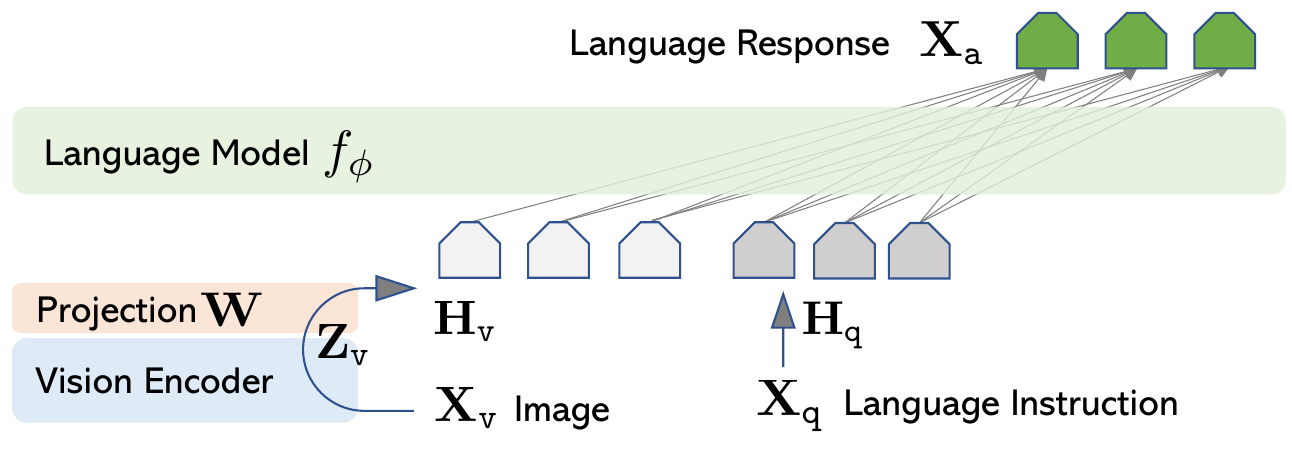}
    \caption{Pipeline of VLM conditional generation in LLaVA~\cite{liu2023llava}.}
    \label{fig:vlm-generative-llava}
\end{figure}

LLaVA incorporates instruction fine-tuning to enhance its multimodal conversational abilities. Liu et al. generated 150k synthetic visual instruction samples to fine-tune the model. The original LLaVA architecture, as shown in Fig.~\ref{fig:vlm-generative-llava}, integrates a pretrained Vicuna language model encoder with a pretrained CLIP ViT-L/14 vision encoder, aligning their outputs into a shared feature space through a linear projector. Its generative objective enables the model to predict each token sequentially, conditioned on previous tokens, thereby facilitating autoregressive response generation. This generative objective could be defined as:
\begin{equation}
    \mathcal{L}_{G}=\sum_{i}\log P(x^i_{a}|x^{i-k}_{a},...,x^{i-1}_{a};\phi),
\end{equation}
where $x^i_a$ denotes the predicted $i^{th}$ word in the answer and $\phi$ represents learnable weights in the large language model.

To leveraging the LLaVA architecture in RS, several pioneering studies have been proposed. For example, GeoChat \cite{kuckreja2024geochat} curates a comprehensive multi-task instruction-following dataset by aggregating various RS datasets, including RSVQAxBEN\cite{lobry2021rsvqa}, FloodNet\cite{rahnemoonfar2021floodnet}, and DOTA\cite{xia2018dota}. After fine-tuning the language model with Low-Rank Adaptation (LoRA) \cite{hu2022lora}, GeoChat demonstrates strong zero-shot performance across a variety of RS tasks.

LHRS-Bot \cite{muhtar2024lhrs} further enhances generative capabilities by progressively fine-tuning both vision and language models using extensive volunteered geographic information and globally available RS images. Targeting refined RS scene comprehension, SkysenseGPT \cite{luo2024skysensegpt} fine-tunes the LLaVA model on a custom dataset to achieve object, regional, and image-level understanding.

Expanding beyond optical imagery, EarthGPT \cite{zhang2024earthgpt} incorporates \acrshort{sar} and infrared images into its instruction-following dataset, improving performance on multi-sensor imagery. For better urban scenario analysis, UrBench \cite{zhou2024urbench} develops multi-view instructions incorporating both RS and street view imagery, highlighting the inconsistent behavior of VLMs when interpreting diverse urban perspectives.

While previous efforts focus on single static image, TEOChat \cite{irvin2024teochat} engages in conversations about \textit{temporal} sequences of earth observation data, including building change and damage assessment, semantic change detection, and temporal scene classification. EarthDial\cite{soni2025earthdial} is a recently proposed conversational assistant for RS, supporting multiple \acrshort{eo} tasks including classification, detection, captioning, VQA, visual reasoning, and grounding, across multi-spectral, multi-temporal, and multi-resolution imagery. It accommodates various modalities, including RGB, SAR, and multispectral bands (e.g., \acrshort{nir} and infrared), and enables bi-temporal and multi-temporal analysis for change detection.

Prior studies often integrate RS imagery with varying \acrshort{gsd}s to accommodate differences in spatial resolution; however, their tuning data is typically limited to low- and mid-resolution images. A distinct challenge in RS lies in interpreting very high-resolution imagery, where image sizes are large and objects of interest are often extremely small. To evaluate the capabilities of existing FMs in this context, Wang et al. \cite{wang2025xlrs} introduced XLRS-Bench, a benchmark constructed from ultra-high-resolution RS imagery, which reveals significant performance limitations of current models under such conditions.

\subsection{Summary and Discussion}

In summary, VLMs in RS aim to capture and model the correlations between various forms of visual data and associated textual descriptions. Research efforts in this area can be broadly categorized into two main directions: contrastive learning, which focuses on aligning image-text pairs through methods such as Visual-Text Models and Visual-Location Models, and generative learning, where the emphasis is on image-conditioned autoregressive text generation, exemplified by LLaVA-based approaches. On one front, larger and more diverse RS datasets are developed to support a range of VLM tasks; on the other, most studies focus primarily on the optical RGB space, with an emphasis on efficiently transferring knowledge from general-domain FMs to RS applications.

While most current VLMs in RS are still tailored to specific applications, they demonstrate early signs of enabling more flexible, scalable, and user-friendly RS solutions. These models offer a novel paradigm for RS interpretation, potentially surpassing traditional, task-specific approaches by fostering more interactive and versatile use of geospatial data.

\section{Other Foundations for RSFMs}\label{sec:other_fms}

In this section, we explore additional types of \acrshort{rsfms} beyond \acrshort{vfms} and \acrshort{vlms}. These include \acrfull{llms} for RS, generative foundation models for RS, and weather forecasting model.

\subsection{Large Language Models for RS}

LLMs, such as the GPT series \cite{radford2018improving,radford2019language,brown2020language,achiam2023gpt}, have become pivotal in advancing AI and FMs across various domains. They have proven to be highly effective to generalize across tasks by leveraging vast amounts of knowledge from large-scale text corpora, often containing billions or trillions of text tokens sourced from the internet.  Despite their widespread success, the application of LLMs in RS remains relatively under-explored, especially in comparison to VFMs and VLMs. This gap is largely due to LLMs lacking the visual perception that is crucial for many RS applications. However, the potential of LLMs to support geospatial prediction tasks through their extensive text-based knowledge remains an valuable yet under-investigated area.

Manvi et al. \cite{manvigeollm} provided early insights, revealing that LLMs such as GPT 3.5 possess an unexpected degree of spatial knowledge. However, they also discovered that simply querying LLMs with geographic coordinates (e.g., latitude and longitude) fails to produce accurate predictions for key geospatial indicators such as population density. To overcome this limitation, they introduced GeoLLM, a method that fine-tunes LLMs using prompts enriched with auxiliary \textit{map} data from OpenStreetMap\footnote{\url{https://www.openstreetmap.org}}. By incorporating spatial context, GeoLLM enables LLMs to more effectively utilize their latent geospatial knowledge. As a result, GeoLLM demonstrated superior performance across multiple critical geo-spatial indicators such as population density, asset wealth, mean income, and women's education.

\subsection{Generative Models and Data Synthesis for RS}

While most previous studies on RSFMs have concentrated on discriminative downstream tasks, generative modeling represents a crucial and complementary direction, particularly for data synthesis. The development of RSFMs requires vast and diverse datasets for effective pretraining; however, RS data is often scarce, incomplete, heterogeneous in modality, and costly to collect or annotate. Synthetic data generation offers a promising solution by expanding both the diversity and quantity of available data, often providing automatic annotations and thereby accelerating the creation of large-scale, versatile RS models.

\textit{Generative foundation models}, especially diffusion models \cite{sohl2015deep,song2021denoising,ho2020denoising,rombach2022high}, have achieved remarkable success in natural image generation, typically leveraging auxiliary information such as text prompts or metadata for controlled generation. Nevertheless, RS data possesses unique characteristics, such as varing spectral bands, irregular temporal sampling, and complex spatio-temporal structures, that present additional challenges beyond those encountered in natural image domains. These differences necessitate the development of specialized generative approaches tailored to RS data properties. Despite these challenges, generative models hold significant potential for advancing RS applications, including super-resolution, cloud removal, temporal inpainting, and cross-modal data generation, underscoring the urgent need for RS-specific generative foundation models.

Recent studies have started to explore this direction. Khanna et al. \cite{khanna2023diffusionsat} introduced DiffusionSat, the first generative FM specifically designed for satellite imagery, inspired by the Stable Diffusion architecture \cite{rombach2022high}. DiffusionSat conditions image generation not only on textual descriptions but also on satellite-specific metadata such as latitude, longitude, timestamps, and \acrshort{gsd}. This metadata-driven conditioning allows the model to be flexibly adapted for diverse generation tasks, including single-image synthesis, super-resolution, inpainting, and temporal generation from publicly available satellite datasets. In parallel, Zheng et al.~\cite{zheng2024changen2} proposed a generative probabilistic change detection model based on diffusion processes, showcasing the utility of generative models for tackling core \acrshort{eo} problems.

In addition to diffusion-based approaches, \textit{multimodal masked modeling} has emerged as another pathway for synthetic data generation. For example, TerraMesh \cite{blumenstiel2025terramesh} constructs a globally diverse multimodal dataset by combining optical imagery, SAR, DEM, and land-cover data, using masked reconstruction objectives to generate synthetic multimodal outputs. This enables the creation of rich, cross-modal training samples that are otherwise difficult and expensive to acquire at large scales.

Another promising direction is \textit{physics-based simulation} for RS data synthesis \cite{xiao2022transfer,songsynrs3d}. Pipelines such as SynRS3D \cite{songsynrs3d} leverage 3D graphics engines like Blender, enhanced with text descriptions generated by GPT-4 and texture realism from Stable Diffusion, to synthesize large-scale, high-resolution optical datasets. SynRS3D features diverse land cover types, precise height information, and annotated building changes, offering a scalable and flexible alternative for creating high-quality RS datasets tailored for training RSFMs.

Collectively, these generative modeling efforts signal a pivotal expansion of RSFM capabilities beyond traditional discriminative tasks, highlighting the importance of synthetic data pipelines in enabling robust, versatile, and scalable remote sensing foundation models.

\subsection{Computational Efficiency and Scalability}

\acrshort{fms} are highly computation-intensive, and RS applications further exacerbate this due to high-resolution, multimodal, and temporally rich data, along with demands for real-time processing and edge deployment. These factors pose substantial challenges to the scalability and efficiency of RSFMs.

A commonly adopted strategy to mitigate training data and computational requirements is the use of \textit{\acrshort{peft} methods} (e.g., LoRA and adapter modules) as well as \textit{knowledge distillation}. As reviewed in previous sections, most of RSFM studies have leveraged these techniques to reduce overhead while maintaining adaptability to RS-specific tasks.

Beyond fine-tuning, another promising direction lies in the design of \textit{lightweight backbone architectures} tailored for RS edge-device deployment. For example, Wang et al. \cite{wang2022unetformer} proposed a streamlined architecture that enhances spatial context modeling in RS imagery by dynamically adjusting the spatial receptive field to accommodate object-scale variations commonly observed in RS data. Similarly, Li et al. \cite{li2024lsknet} introduced a dual-branch CNN–Transformer hybrid architecture optimized for on-orbit inference, striking a balance between representational capacity and computational efficiency.

In addition, \textit{model compression} remains a critical yet relatively underexplored area in RSFMs. Its primary goal is to reduce model size, inference latency, and memory footprint while preserving task performance, which is an essential step for enabling deployment in resource-constrained environments. An early effort in this direction was introduced by Shinde et al. \cite{shinde2025model}.

Given the multimodal heterogeneity and spatiotemporal complexity of RS data, future efforts should target specialized strategies, including multimodal-aware compression, adaptive inference, and dynamic resource management, to enable practical and scalable RSFM deployment.

\subsection{Weather Forecasting Foundation Models}
RSFMs have also made impressive advancements in other geoscience applications, particularly in weather forecasting. A prime example is \textit{Pangu-Weather} \cite{bi2023accurate}, an AI-driven system that generates highly accurate deterministic forecasts. Trained on 39 years of global data, Pangu-Weather leverages a custom three-dimensional Earth-specific transformer (3DEST) architecture, which incorporates Earth-related priors like height into the model. Additionally, it employs a hierarchical temporal aggregation algorithm, where models are trained for progressively longer forecast lead times. Compared to the world’s top traditional numerical weather prediction (NWP) systems, Pangu-Weather consistently outperforms, underscoring the transformative potential of RSFMs in \acrshort{eo} and related geoscience applications.

\subsection{Responsible RSFMs}

As RSFMs continue to advance, it is essential to consider their broader impacts beyond technical innovation. The large-scale training and deployment of RSFMs raise important environmental and ethical concerns, while also offering the potential for substantial societal benefit \cite{bommasani2021opportunities}. This is particularly relevant given the strategic role of RS in \acrshort{eo}, environmental monitoring, disaster response, and national security.  A recent effort by Ghamisi et al. \cite{ghamisi2025responsible} provides a timely and comprehensive review on responsible AI for \acrshort{eo}. Their work highlights key dimensions including security, privacy, ethical risks, algorithmic bias, and fairness, and proposes achievable pathways to align model development with societal values and public interest. While our survey emphasizes the technical foundations of RSFMs, we refer readers to this work for an in-depth discussion of responsible practices and their relevance to the RS community.

\subsection{Summary and Discussion}
In summary, while LLMs and generative FMs from general domains have demonstrated revolutionary performance across various applications, their counterparts in RS still trail behind. However, recent progress in adapting these models for RS has been promising, revealing significant potential and opportunities. Additionally, specialized RS applications, such as weather forecasting, highlight the need for tailored RSFMs that integrate advanced AI techniques with domain-specific RS knowledge. This underscores the growing importance of deeper integration between AI innovations and geospatial science to fully harness the capabilities of RSFMs.

\section{Benchmark Performance}\label{sec:benchmarks}

In this section, we compare, analyze, and discuss the existing RSFM studies reviewed in this survey. It is important to note that these studies have distinct research focuses, modalities, learning setups, and target tasks or applications, making it difficult to conduct fair comparisons for all of them using unified benchmarks. Furthermore, as RSFM development is still in its early stages, robust and comprehensive evaluation frameworks across different tasks and modalities are yet to be fully established.
Given these limitations, we focus on key dimensions and try to provide an informative analysis and insights into the current progress and future directions.

\begin{table*}[t]
    \scriptsize
    \setlength\tabcolsep{8.2pt}
    \caption{Performance of commonly used downstream benchmarks by different pre-training methods. *, $^{\star}$, $^{\dagger}$ denotes that metric numbers are sourced from \cite{guo2024skysense}, \cite{li2024masked}, \cite{noman2024rethinking}, respectively. "FT" denotes fine-tuning.}
    \centering
    \begin{tabular}{|l|c|cc|c|ccc|c|c|c|c|}
    \hline
        Method & Backbone & \multicolumn{2}{c|}{BigEarthNet (mAP)} & EuroSAT & \multicolumn{3}{c|}{Onera Satellite} & SpaceNet & fMoW RGB & DIOR-H & DIOR-R \\
        & & 10\% FT & 100\% FT & Acc. & Prec. & Rec. & F1 & mIoU & Acc. & mAP$_{50}$ & mAP\\
    \hline
       SeCo\cite{manas2021seasonal} & ResNet-18 & 81.9 & 87.3 & 93.1 & 65.5 & 38.1 & 46.9 & - & - & - & - \\
       SeCo\cite{manas2021seasonal} & ResNet-50 & 82.6 & 87.8 & - & - & - & - & - & - & - & - \\
       GASSL\cite{ayush2021geography} &  ResNet-50 & 80.2 & - & 89.5$^{\dagger}$ & - & - & 46.3* & 78.5 & 71.6 & 67.4* & 65.7*\\
       MATTER\cite{akiva2022self} & ResNet-34 &  - & 88.0 & - & 61.8 & 57.1 & 59.4 & 81.1 & - & - & - \\
       SatMAE\cite{cong2022satmae} & ViT-L & 82.1 & - & 98.9 & - & - & 52.8* & 78.1 & 77.8 & 70.9* & 65.7* \\
       CACo\cite{mall2023change} & ResNet-18 & - & - & - & 60.7 & 42.9 & 50.3 & - & - & - & - \\
       CACo\cite{mall2023change} & ResNet-50 & 81.3* & 87.0* & - & 62.9 & 44.5 & 52.1 & - & - & 66.9* & 64.1* \\
       GFM\cite{mendieta2023towards} & Swin-B & 86.3 & - & - & 58.1 & 61.7 & 59.8 & - & - & 72.8* & 67.7* \\
       Scale-MAE\cite{reed2023scale} & ViT-L & - & - & - & - & - & - & 78.9 & 77.9 & 73.8* & 66.5* \\
       CSP\cite{mai2023csp} & ResNet-50 & - & - & - & - & - & - & - & 71.0 & - & - \\
       CROMA\cite{fuller2024croma} & ViT-B & 85.0 & - & 99.2 & - & - & - & - & - & - & - \\
       CROMA\cite{fuller2024croma} & ViT-L & 85.0 & - & 99.5 & - & - & - & - & - & - & - \\
       SkySense\cite{guo2024skysense} & ViT-L & 88.7 & 92.1 & - & - & - & 60.1 & - & - & 78.7 & 74.3 \\ 
       msGFM\cite{han2024bridging} & Swin-B & 87.5 & 92.9 & - & - & - & - & - & - & - & - \\
       SatMAE++\cite{noman2024rethinking} & ViT-L & 85.1 & - & 99.0 & - & - & - & - & 78.1 & - & -\\
       MA3E\cite{li2024masked} & ViT-B & - & - & - & - & - & - & - & - & - & 71.8 \\
    \hline
    \end{tabular}
    \label{tab:benchmark-ssl}
\end{table*}

\begin{table*}[t]
    \centering
    \caption{Zero-shot performance comparison of VLMs on RS image classification and retrieval benchmarks with Sentinel-2 data.}
    \label{tab:clip-sota}
    \setlength\tabcolsep{13.5pt}
    \scriptsize
    \begin{tabular}{|l|c|c|cc|cccc|}
    \hline
        \multirow{2}{*}{Model} & Downstream & \multirow{2}{*}{Backbone} & \multicolumn{2}{c|}{Classification} & \multicolumn{4}{c|}{Retrieval} \\
        & Annotations & & EuroSAT & BEN & \multicolumn{2}{c}{EuroSAT} & \multicolumn{2}{c|}{BEN} \\
        & & & Acc. & mAP & mAP$^{100}$ & mAP$^{20}$ & mAP$^{100}$ & mAP$^{20}$ \\
    \hline
        CLIP~\cite{radford2021learning} & \ding{55} & ViT-B/32 & 47.61 & 21.31 & 46.86 & 49.61 & 30.45 & 32.20 \\
        CLIP~\cite{radford2021learning} & \ding{55} & ViT-B/16 & 53.59 & 23.13 & 63.99 & 72.57 & 32.54 & 33.10 \\
        CLIP-RSICD\footnotemark & \checkmark & ViT-B/32 & 45.93 & 27.68 & 49.88 & 53.50 & 38.76 & 36.11 \\
        RemoteCLIP~\cite{liu2024remoteclip} & \checkmark & ViT-B/32 & 38.59 & 22.76 & 51.21 & 53.27 & 34.39 & 38.42 \\
        GRAFT~\cite{mall2023remote} & \ding{55} & ViT-B/32 & 47.80 & 29.31 & 66.12 & 72.36 & 45.88 & 45.35 \\
        GRAFT~\cite{mall2023remote} & \ding{55} & ViT-B/16 & 63.76 & 32.46 & 81.56 & 85.21 & 49.61 & 53.86 \\
    \hline
    \end{tabular}
\end{table*}

\subsection{Performance of VFM Pre-training}

As outlined in Section~\ref{sec.sup}, VFM pre-training models are subsequently fine-tuned for various downstream RS tasks~\cite{lacoste2024geo}. 
Table \ref{tab:benchmark-ssl} presents evaluations for seven widely adopted downstream tasks across different modalities, including: BigEarthNet~\cite{sumbul2019bigearthnet} for \textit{multi-label land cover classification}; EuroSAT~\cite{helber2019eurosat} for \textit{\acrshort{lulc} classification}; Onera Satellite~\cite{daudt2018urban} for \textit{change detection} tasks; SpaceNet~\cite{van2018spacenet} for \textit{building segmentation}; Functional Map of the World (FMoW)~\cite{christie2018functional} for \textit{high-resolution satellite image time series classification} among 62 categories;  DIOR-H~\cite{li2020object} for \textit{horizontal object detection} and DIOR-R~\cite{cheng2022anchor} for \textit{oriented object detection}.
All metrics are sourced from the respective papers. We select the most widely used datasets for this comparison, though other benchmarks, such as Geo-Bench~\cite{lacoste2023geo}, along with those tailored for different tasks, could also provide valuable insights.

From the table, we can observe that diverse RS tasks benefit from increasingly advanced pre-training methods. Additionally, many pre-training approaches show effectiveness across multiple downstream tasks. However, it is essential to recognize that different methods may target specific applications, underscoring the need for comprehensive and standardized benchmarks for fair comparison. Additionally, there remains considerable potential for development in this field, indicating significant opportunities for future research.

\footnotetext{\url{https://github.com/arampacha/CLIP-rsicd}}

\subsection{Performance of VLM Transfer Learning}

% This subsection compares the zero-shot performance of various VLMs in RS. We first source benchmark results from \cite{mall2023remote} as in Table~\ref{tab:clip-sota}, which evaluated different types of FMs, including general-domain VLMs like CLIP and RS-specific models fine-tuned from CLIP, such as CLIP-RSID\footnotemark[2], RemoteCLIP\cite{liu2024remoteclip}, and GRAFT\cite{mall2023remote}, across typical RS tasks like image classification and retrieval. 
This subsection evaluates the zero-shot performance of various VLMs in RS. We begin with a \textit{closed-set} evaluation, comparing general-domain VLMs like CLIP with RS-specific models fine-tuned from CLIP, such as CLIP-RSID\footnotemark[2], RemoteCLIP~\cite{liu2024remoteclip}, and GRAFT~\cite{mall2023remote}, on key RS tasks like image classification and retrieval.
These models enable zero-shot evaluation by projecting images and text into a shared embedding space. The process involves extracting image features via an image encoder and generating text embeddings from class-specific prompts (e.g., arbitrary category names) using a text encoder. The model then calculates cosine similarity between the image and text embeddings, assigning the label with the highest similarity score. Performance is measured using standard classification metrics, allowing models to generalize across diverse tasks without additional task-specific training.

Table~\ref{tab:clip-sota} present benchmark results sourced from \cite{mall2023remote}.  It shows that while CLIP demonstrates some effectiveness on RS datasets, its performance falls short compared to its success on natural image benchmarks. In contrast, recent RS-specific VLMs exhibit significant improvements, underscoring the potential of tailoring VLMs for RS applications. However, these specialized models tend to perform well on specific benchmarks but show less overall robustness compared to CLIP. These results highlight both the progress made and the substantial opportunities for future advancements in this domain.

Another type of benchmarks base on \acrshort{vqa} for \textit{open-set} evaluation. These models perform zero-shot evaluation by encoding images and questions separately, then fusing their representations to understand the relationship between visual and textual information. Based on this joint representation, different model types either generates an answer in an open-ended format or selects the most probable answer from a predefined set.

\begin{figure}[t]
    \centering
    \includegraphics[width=\linewidth]{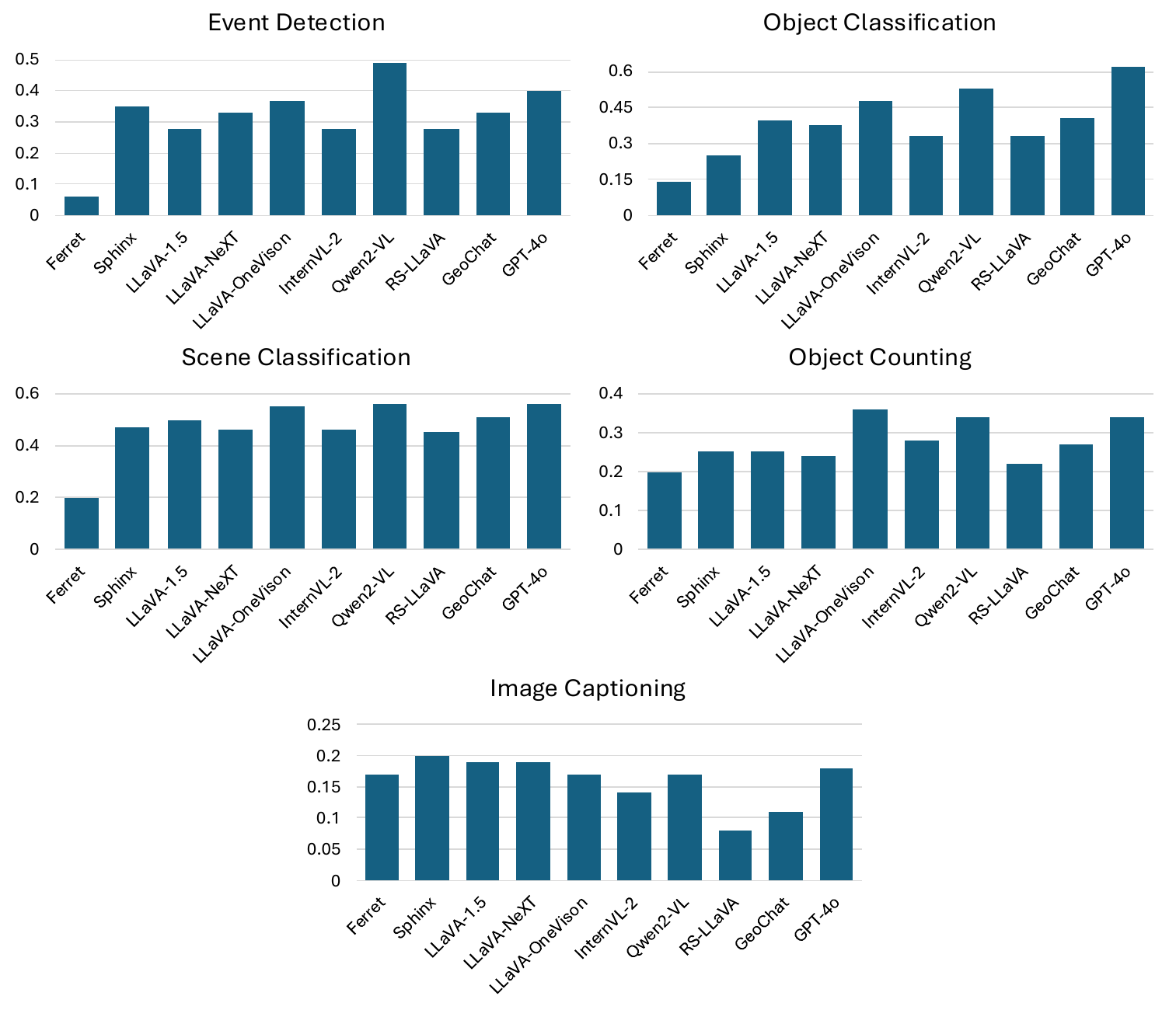}
    \caption{Zero-shot performance comparison of typical VLMs across different tasks on GEOBench-VLM\cite{danish2024geobench}.}
    \label{fig:bench_geobench_vlm}
\end{figure}

As an example of \textit{open-set} benchmark, GEOBench-VLM~\cite{danish2024geobench} was recently introduced for evaluating VLMs in RS, encompassing 31 fine-grained tasks across eight broad categories, including scene and object classification, object detection, segmentation, captioning, event detection, and non-optical and temporal understanding tasks. This benchmark provides a comprehensive assessment of perception tasks for VLMs in geospatial applications. Figure~\ref{fig:bench_geobench_vlm}, presents evaluations of 10 open-source and advanced proprietary VLMs. The results reveal a significant performance gap between open-source and closed-source models, while all models exhibit suboptimal performance across various geospatial recognition tasks. The primary limitation stems from data biases in training these models, as current large-scale multimodal datasets contain limited geospatial data, restricting their ability to generalize effectively to RS-specific tasks. This highlights the need for more advanced and specialized models to support complex Earth observation and geospatial interpretation tasks, underscoring the importance of dedicated geoscience and remote sensing efforts in building more powerful and domain-adapted RSFMs to bridge this gap.

\section{Future Directions}\label{sec.future}

Although significant progress has been made in RSFMs, they are still in their early stages, especially compared to their counterparts in general foundation models. However, the potential and opportunities for growth in this field are immense. Below, we outline several key future directions.

\textit{Larger and More Diverse Datasets}: FMs thrive on vast amounts of data, with the principle being "the more data, the better" \cite{bommasani2021opportunities}. However, the current RS datasets fall short of supporting the training of truly large-scale RSFMs. A notable observation is the absence of the “\textit{emergent capabilities}” observed in general-domain FMs, indicating that the existing scale and diversity of RS data are insufficient to enable comparable advances. In parallel, although RSFMs fine-tuned from general FMs often yield superior performance on specific benchmarks, they still fall short in generalizing to out-of-distribution data.

Future dataset construction efforts should not focus solely on increasing data volume, but also prioritize critical dimensions such as modality diversity, spectral and temporal resolution, and dataset biases. As summarized in Tables \ref{tab. pre-train-datasets} and \ref{tab:VLM-RSDataset}, most available datasets are still centered on optical RGB and MSI. While there is some variation in \acrshort{gsd}, the majority of studies rely on mid-resolution, publicly available sources like Sentinel imagery. To support more complex geospatial tasks, there is a pressing need for higher-resolution datasets with richer contextual information. Additionally, although recent initiatives have started to incorporate temporal sequences for RS change detection, significant progress is still needed. 

Moreover, annotating large-scale, multimodal RS datasets remains a major bottleneck due to the inherent complexity and heterogeneity of the data. As the scale of RS data continues to grow, robust data management practices, including integration, quality assurance, governance, and privacy controls, will become increasingly critical. Despite rapid advancements, we anticipate that the next stage of RSFM development will require datasets that are not only larger and more diverse, but also more structurally and semantically enriched.

\textit{Multimodal RSFM}: An ideal RS interpretation system should be capable of processing multiple sensor modalities simultaneously, providing a richer and more holistic view of geospatial data. Beyond sensor modalities, RSFMs should also incorporate other data types, such as text, dense annotations for segmentation, and bounding boxes for object detection, enabling them to handle a broad spectrum of tasks with greater flexibility.

\textit{Spatiotemporal Processing}: Unlike natural images, RS imagery is often irregularly sampled across time and varies in scale and resolution due to differences in sensors and platforms. Developing RSFMs with advanced spatiotemporal processing capabilities will be critical for observing and analyzing dynamic changes in the Earth's surface, enhancing both short-term monitoring and long-term environmental analysis.

% \textit{High Processing Speed}: In time-sensitive \acrshort{eo} applications such as disaster monitoring, real-time data processing is crucial for effective response and decision-making. RSFMs shall be designed to handle vast amounts of RS data with high efficiency, enabling rapid analysis and forecasting. This will ensure that RSFMs can provide actionable insights swiftly, supporting critical tasks like early warning systems, damage assessment, and emergency response coordination.

\textit{Efficient transfer}: While general-domain FMs offer powerful capabilities for processing natural data, adapting these models for RS applications is both a cost-effective and practical solution. Adaptation typically involves conditioning the foundation model with new information, either by incorporating additional data or prompts into its input or by selectively updating parts of the model’s parameters to align with the new context. To achieve efficient transfer, advanced techniques like \textit{task specialization}, \textit{spatiotemporal adaptation}, and \textit{domain-specific tuning} are essential. Furthermore, \textit{continuous learning} approaches should be explored to allow models to evolve with new data while avoiding catastrophic forgetting, ensuring sustained performance and relevance for RS tasks.

\textit{Scalability and Computational Efficiency}: The scalability of RSFMs remains a key challenge due to the high computational cost associated with processing large, multi-modal, and high-resolution RS data. Compared to conventional vision tasks, RS applications often demand greater memory and processing power, particularly for multi-temporal and hyperspectral analysis. These challenges are amplified in scenarios requiring on-board processing on satellites, UAVs, or edge devices, where computational resources are limited. Despite their importance, few studies have explored model compression techniques such as pruning, quantization, or knowledge distillation, for tailoring large RSFMs to lightweight, efficient variants. Such strategies are critical to improving adaptability and deployment feasibility. In time-sensitive EO applications like disaster response and environmental monitoring, RSFMs must support real-time or near-real-time inference. Future work should prioritize optimizing RSFMs to balance performance with efficiency, enabling timely, scalable, and actionable geospatial intelligence.

\textit{RSFMs for broader EO applications}: The unique nature of RS data, combined with the diverse range of EO applications, extends well beyond the typical tasks handled by general-domain FMs. Developing RSFMs that support a broader spectrum of EO applications—such as weather forecasting, environmental monitoring, and urban planning—is a highly valuable yet challenging task. These RSFMs need to process a wide variety of input modalities, such as satellite/airborne imagery, GPS signals, temperature data, and RS-derived products like Normalized difference vegetation index (NDVI) \cite{rouse1974monitoring} maps, while delivering outputs that can range from atmospheric predictions to land-use assessments. Achieving this requires not only advanced AI methodologies but also deep expertise in RS and geoscience, reflecting the complex, interdisciplinary nature of these applications.

\textit{Reasoning for Geospatial Tasks}: Most existing RSFMs primarily focus on perception-oriented tasks such as scene classification, object detection, and semantic segmentation, with only limited extensions to relatively simple reasoning functions (e.g., spatial relation judgment). These capabilities form the foundation of many geospatial applications. However, real-world \acrshort{eo} tasks often require higher-order reasoning that goes beyond direct recognition. For example, analysts may need to integrate multi-temporal observations, infer causal relationships (e.g., attributing deforestation to specific policy or climate drivers), or correlate multimodal data sources to detect abnormal patterns or changes. These tasks demand \textit{spatial reasoning} (e.g., understanding topological relationships), \textit{temporal reasoning} (e.g., trend forecasting or event progression), and \textit{causal reasoning} (e.g., identifying cause-effect dynamics in land use change). Unlike general FMs, which typically focus on linguistic or commonsense reasoning in structured or textual contexts, geospatial reasoning must operate over noisy, high-dimensional, and sparsely labeled spatial data. Advancing RSFMs to support such reasoning capabilities will require integrating domain knowledge, incorporating causal and temporal models, and developing architectures capable of synergistic data fusion across modalities and scales. Such models would not only perceive but also explain, predict, and reason over complex EO scenarios, enabling next-generation geospatial intelligence systems.

\textit{Explainability of RSFMs}: As RSFMs are increasingly adopted in critical EO applications, enhancing their interpretability becomes essential for ensuring trust, transparency, and accountability. The black-box nature of large models poses challenges for understanding how decisions are made, particularly when integrating heterogeneous sensor modality inputs beyond optical RGB. While explainable AI (XAI) has made progress in general vision tasks, its application in RS remains limited. Developing modality-aware explainability tools tailored to RS data characteristics is a meaningful direction, enabling users to trace predictions back to geospatial and physical evidence. This will be especially crucial in sensitive applications such as environmental policy, disaster assessment, and climate monitoring.

\textit{Generalization of RSFMs}: A key requirement for RSFMs is the ability to generalize across diverse spatial, temporal, and sensor domains. However, RS data often exhibits substantial distribution shifts due to seasonal changes, geographic variability, and differing acquisition conditions. Many current RSFMs still rely on fine-tuning with task-specific data, which limits their generalization capabilities. Future work should investigate domain-agnostic pretraining objectives, robust alignment strategies across modalities, and data augmentation techniques that simulate real-world variability. Improving out-of-distribution generalization will be critical for deploying RSFMs in operational settings where annotated data is scarce or unseen conditions prevail.

\textit{Mixture of Experts}: The Mixture of Experts (MoE) approach offers a promising way to enhance both the adaptability and efficiency of RSFMs by leveraging specialized sub-models (or experts) tailored to handle different tasks or data modalities. Given the large diversity of RS data and the complexity of EO applications, building a single unified RSFM can be challenging. An MoE framework addresses this by dynamically selecting or combining experts that are specialized for specific RS tasks. This method not only improves task-specific predictions but also enhances computational efficiency by activating only the required experts for each task, thus reducing unnecessary processing. Moreover, integrating MoE into RSFMs enables a more flexible system, capable of scaling across various geospatial applications without the need for extensive retraining. However, designing efficient routing mechanisms to allocate tasks to the appropriate experts while ensuring seamless coordination between them remains a significant challenge.

\section{Conclusion}
In this survey, we have provided a comprehensive review of the latest advancements in Remote Sensing Foundation Models, exploring their background, foundational concepts, datasets, technical approaches, benchmarking efforts, and future research directions. Our survey aims to offer a clear and cohesive overview of the current developments in RSFMs, serving as a valuable resource for guiding future research in this rapidly growing field. We hope this work will not only help to catalog future studies but also foster a deeper understanding of the challenges that remain in building intelligent Earth observation systems and advancing remote sensing interpretation.

\section*{Acknowledgments}
This work was supported in part by the Council for Science, Technology and Innovation (CSTI), the Cross-ministerial Strategic Innovation Promotion Program (SIP), Development of a Resilient Smart Network System against Natural Disasters (Funding agency: NIED), JST, FOREST under Grant Number JPMJFR206S. It was also partly supported by the Ministry of Education Singapore through the Tier-2 Scheme under Project MOE-T2EP20123-0003.

\appendix[Figure Credits]
{
\scriptsize
\noindent\textbf{Fig.~\ref{fig:rs-modalities}: Example of different RS modalities.}

\noindent \textit{Optical RGB}: 
Source image from \url{https://maps.gsi.go.jp/development/ichiran.html}

\noindent \textit{HSI}: 
Source image from \url{https://commons.wikimedia.org/w/index.php?curid=25442380}

\noindent \textit{LiDAR}: 
Source image from \url{10.1109/MGRS.2019.2893783}

\noindent \textit{TIR}: 
Source image from \url{https://www.esa.int/ESA_Multimedia/Images/2022/07/Land-surface_temperature_in_Prague_on_18_June_2022}

\noindent \textit{DSM}: 
Source image from \url{https://commons.wikimedia.org/w/index.php?curid=24447099}

\noindent \textit{SAR}: 
Source image from \url{https://commons.wikimedia.org/w/index.php?curid=117320}}

\printglossary[type=\acronymtype]
{\small
\bibliographystyle{IEEEtran}
\bibliography{main}
}
\end{document}